\documentclass{sfbart}

\usepackage{chicago,ecltree,epic,eepic}
\allowdisplaybreaks

\newcommand{\Reals}{\mathrm{I} \! \mathrm{R}}
\newcommand{\Rn}{{\mathrm{I} \! \mathrm{R}}^n}
\newcommand{\Nats}{\mathrm{I} \! \mathrm{N}}

\newcommand{\Xsf}[1]{\mbox{\sf #1}}

\newcommand{\Po}{\mbox{$\mathcal{P} \:$}}

\newcommand{\Rl}{\mbox{$\mathcal{R(L)} \:$}}
\newcommand{\La}{\mbox{$ \mathcal{L}\:$}}

\newcommand{\R}{\mbox{$ \mathcal{R}\:$}}

\newcommand{\gr}{\mbox{$\stackrel{r}{\longrightarrow}\:$}}
\newcommand{\cs}{\mbox{$\stackrel{c}{\longrightarrow}\:$}}

\newtheorem{de}{Definition}
\newtheorem{po}{Proposition}

\begin{document}

\def\copyrightyearmodC{97}
\def\currentmonth{Oktober}
\def\currentissue{117}
\def\cpright{Der Autor}

\title{Probabilistic Constraint Logic Programming}

\author{Stefan Riezler}

\address{Graduiertenkolleg ILS \\
Seminar f\"ur Sprachwissenschaft \\
Universit\"at T\"ubingen \\
Wilhelmstr. 113 \\
72074 T\"ubingen \\
Germany}

\email{riezler@sfs.nphil.uni-tuebingen.de}

\thanks{This work was supported by the
  Graduiertenkolleg Integriertes Linguistikstudium at the University
  of T\"ubingen and by the Teilprojekt B7 of the
  Sonderforschungsbereich 340 of the Deutsche Forschungsgemeinschaft.}
\thanks{I am greatly indebted to Steven Abney for coaching this work
  in many helpful discussions. Furthermore, I would like to thank
  Graham Katz and Detlef Prescher for their valuable comments.}

\begin{abstract}
This paper addresses two central problems for probabilistic processing
models: parameter estimation from incomplete data and efficient
retrieval of most probable analyses. These questions have been answered
satisfactorily only for probabilistic regular and context-free
models. We address these problems for a more expressive
probabilistic constraint logic programming model.

We present a log-linear probability model for probabilistic constraint
logic programming. On top of this model we define an algorithm to
estimate the parameters and to select the properties of log-linear
models from incomplete data. This algorithm is an extension of the
improved iterative scaling algorithm of
\citeN{Della-Pietra:95}. Our algorithm applies to log-linear models in
general and is accompanied with suitable approximation methods when
applied to large data spaces. Furthermore, we present an approach for searching
for most probable analyses of the probabilistic constraint logic
programming model. This method can be applied to the ambiguity resolution
problem in natural language processing applications.
\end{abstract}

\maketitle

\section{Introduction}

\citeN{Rabiner:89} identified three basic problems of
interest that must be solved for a Hidden Markov Model 
to be useful in real-world speech recognition applications: the
parameter estimation problem, the optimal state sequence problem and
the observation sequence probability problem. These problems 
generalize to arbitrary probabilistic symbol processing models in various
real-world applications in an obvious manner.  The first two problems can
be stated in a more general way as follows.

\begin{enumerate}
\item Let an unanalyzed observation sequence $O = O_1, \ldots O_n$ and a probabilistic processing model with parameter set $\lambda$ be
  given, and suppose that the value of $\lambda$ is unknown and $O$
  forms a random sample from the distribution involving $\lambda$, how
  can the model parameters $\lambda$  be estimated? 
\item Given $O_i$ and $\lambda$, how can the most probable analysis of
  the input $O_i$ be found efficiently?
\end{enumerate}

Recent interest in probabilistic models of natural language processing
can be attributed to the fact that solutions to the above-mentioned
general problems can lead quite directly to effective, but
conceptually simple and mathematically clear solutions to various
problems in natural language processing.

This connection can be
illustrated with the problem of ambiguity resolution
(or disambiguation or parse ranking) as follows: Grammars
describing a nontrivial fragment of natural language may attach a
large number of different analyses to sentences of reasonable length. Since not
all of these  analyses are in accord with human perceptions, there is
clearly a need to distinguish more plausible analyses of an input from
less plausible or totally spurious ones. The simple but effective idea
adopted in probabilistic grammars is to connect the plausibility of an
analysis with its probability. In this vein the correct, i.e., most
plausible analysis of a string is assumed to be the most probable
analysis of the string. A solution to problem 1 will adapt the model
parameters $\lambda$ to the input corpus $O$ and thus justify the
assumption that the correct parse of a string $O_i$ is the
most probable parse of $O_i$ as produced by the grammar parametrized
by $\lambda$. A solution to problem 2 will yield an algorithm to
search for the most probable parse of a given input string $O_i$ as
produced by a probabilistic grammar with parameter set $\lambda$.

Most popular approaches to solving these problems in the area of
natural language processing are based on Baum's maximization
technique, which is known as the ``Baum-Welch algorithm''
\cite{Baum:67,Baum:70,Baum:72}.
This algorithm estimates the parameters of a
Hidden Markov Model, i.e., a stochastic regular grammar, in a framework
of maximum likelihood estimation from incomplete data. This means, the
parameters are iteratively reestimated until convergence to a set of values which
locally maximize the likelihood function, i.e., the probability that the
model assigns to the given unanalyzed observation sequence. In this
sense the model parameters are adjusted to best describe a given observation
sequence. The estimation algorithm can be defined inductively forwards
and backwards, yielding the efficient ``forward-backward
algorithm''\footnote{See \citeN{Rabiner:89} for a tutorial.}.
\citeN{Baker:79} generalized this algorithm to the so-called ``inside-outside
algorithm''\footnote{See \citeN{Lari:90} and
\citeN{Jelinek:90} for introductions.},
which efficiently estimates the parameters of a stochastic
context-free grammars.
Both algorithms are special instances of the EM-algorithm for
maximum-likelihood estimation from incomplete data \cite{Dempster:77}.
A dynamic-programming approach similar to the one used in the
efficient versions of the parameter estimation algorithms can be used
to find the most probable analysis of stochastic context-free and
stochastic regular grammars and is known as the ``Viterbi-algorithm''
\cite{Viterbi:67}. 

The class of algorithms based upon Baum's maximization technique
includes not only regular and context-free versions but recently has
been extended to, e.g., stochastic context-free grammars with
bracketing constraints \cite{Pereira:92} and feature-based constraints
\cite{Briscoe:92}, stochastic depencendy grammars
\cite{Carroll:92} and stochastic lexicalised tree-adjoining grammars 
\cite{Resnik:92,Schabes:92}.
Despite the generality of the algorithm, there are clear restrictions
on the expressivity of the probabilistic processing models the
algorithm can be applied to. Even if the structural operations of the
probabilistic processing model may be sensitive to contextual
features, this context-sensitivity has to be internal to the
structural elements combined. The combination process itself has to be
context-free, i.e., in terms of probability theory, different
stochastic derivation choices at the same time-step have to be
independent of the history of the derivation process and also
independent of one another. 

This fact poses a problem for attempts to build stochastic
versions of grammars which are more expressive than context-free. The
grammars we are interested in here are constraint logic grammars (CLGs), i.e.,
highly expressive constraint-based grammars formalized in a
(Turing-)powerful framework of constraint logic programming
(CLP)\footnote{CLP provides one possible approach to an operational
  treatment of various purely declarative grammar frameworks by an
  embedding of arbitrary logical languages into constraint logic
  programs. CLGs thus are simply understood as grammars formulated by
  means of a suitable logical language which can be embedded as a
  constraint language into a CLP scheme. Examples for an embedding of
  feature-based logical languages into the CLP scheme of
  \citeN{HuS:88} are the approaches of \citeN{Doerre:93} and
  \citeN{Goetz:95}.}.
This treatment of CLGs as special applications of CLP will allow us to
refer in the following to the general framework of CLP.
Stochastic versions of CLP exhibit a context-sensitivity problem in
that incompatible variable bindings can lead to failure derivations in
dependence of the (simultaneous) history of the derivation
process.
\citeN{Eisele:94}, \citeN{Miyata:96} and \citeN{Osborne:97},
who attempt to adapt Baum's maximization technique to estimate the
parameters of their stochastic constraint-based models, try to escape from this
problem by redefining the derivation process of their respective
probabilistic processing model to include only successful derivations
and by renormalizing the probability distribution over derivations.
Unfortunately, this move contradicts the basic independency
assumption made in the parameter estimation algorithm and prohibits an
application of Baum's technique as an optimization
algorithm in maximum likelihood estimation of stochastic CLP.

To date to our knowledge there is no approach which solves problems 1
and 2 satisfactorily for a probabilistic model of CLP.
However, an excellent starting point is the approach to
``stochastic attribute-value grammars'' of \citeN{AbneySAVG:96}. Abney
presents a probabilistic model of grammars which produce analyses in form
of dags (directed acyclic graphs) by defining the probability
distribution over these dags as a random field. For such probability
models algorithms to estimate parameters from complete data exist
\cite{Della-Pietra:95} and are shown to be applicable to the
stochastic grammar model. However, complete data means large
corpora of costly manually analyzed, i.e., hand-parsed, data. So one
open question is how to estimate parameters from incomplete, unanalyzed
input, i.e., from simple corpora of natural language strings. Furthermore, if
the intended application is ambiguity resolution, a second question is
how to use the structure of the probabilistic model to guide the
search for the most probable analysis of a string rather than simply
listing all possible analyses and choosing the best one.

The aim of this paper is to present a
probabilistic model of CLP and to couple this with an algorithm to induce
the parameters and properties of such models from incomplete data and
with an algorithm to search for best analyses.
Our approach to probabilistic CLP is based on a log-linear probability
model, i.e., a powerful exponential probability model well-known in
probabilistic network modeling 
\cite{Geman:84,Ackley:85,Pearl:88}.
On top of this probability model we define an algorithm to estimate
the parameters and to select the properties of log-linear models from
incomplete data. This induction algorithm is an extension of the
improved iterative scaling algorithm of \citeN{Della-Pietra:95}
adjusted to incomplete data. The techniques developed in this context
apply to log-linear models in general and are accompanied with
suitable Monte Carlo approximations when applied to large data spaces.
For the intended CLP application we build upon the CLP scheme of
\citeN{HuS:88}. In this context we present an algorithm to search for
most probable analyses of the probabilistic CLP model. This algorithm
is formulated as a probabilistic version of Earley deduction and can
be applied to the ambiguity resolution problem in natural language
processing applications\footnote{
Even if a solution to our two problems
can be seen as a necessary prerequisite for further applications such
as grammar induction or language modelling, it is a necessary
\emph{and} sufficient prerequisite only for the application of
ambiguity resolution. For the application of grammar induction, the
question of how to impose useful constraints on the form of possible
analyses in order to reduce the number of parameters to be estimated will
become important. In language modelling applicatons, a shift of focus
from imposing a probability distribution over a given set of analyses
to imposing a probability distribution over input strings is made.}.

The remainder of this paper is organized as follows. 
Section \ref{CLP} introduces the basic formal concepts of CLP.
Section \ref{Baum} discusses in more detail the above-mentioned
context-sensitivity problem in case of parameter estimation from
incomplete data. 
Section \ref{LogLinearModel} presents a general log-linear model for
probabilistic CLP.
Problem 1, i.e., parameter estimation of log-linear probability models from
incomplete data is treated in Sect. \ref{Induction}. An algorithm for
automatically selecting properties of log-linear
models in the presence of incomplete data is also presented. 
Section \ref{Approximation} discusses the problem of estimating the
terms in the formula presented in Sect. \ref{Induction} in the
presence of large sample spaces by Monte Carlo methods.
Problem 2, i.e., methods to search for most probable analyses for
probabilistic CLP, is approached in Sect. \ref{Search} in the
form of a probabilistic version of Earley deduction.
Section \ref{Hugh} gives some concluding remarks and discusses the relation of
probabilistic CLP to other probabilistic processing models.

\section{Constraint Logic Programming} \label{CLP}

In the following we will quickly report the basic definitions of 
the CLP scheme of \citeN{HuS:88}. This scheme is a powerful extension
of conventional logic programming (see \citeN{Lloyd:87}) and also of
the CLP scheme of \citeN{JuL:86} by an
incorporation of arbitrary constraint languages and corresponding
constraint solving methods into logic programming languages.

For example, Prolog is obtained by employing equations between first order
terms as constraint language and by interpreting these equations in the
Herbrand universe. The corresponing operational semantics
of SLD-resolution can be seen to rely on a constraint solver which
solves term equations in the Herbrand universe by term unification.

A constraint logic program \Po is then defined with respect to an implicit
basic constraint language \La and its relational extension \Rl as
follows (see \citeN{HuS:88}).

\begin{de}[definite clause specification]
A definite clause specification \Po in \Rl is a set of
definite clauses of the form
\begin{center}
$A \leftarrow B_1 \: \& \ldots \& \: B_n \: \& \: \phi$ \end{center}
where $A, B_1, \ldots, B_n$ are \Rl-atoms, 
$r(\vec{x}) $ is an \Rl-atom iff $r \in \R$ is a relational symbol with arity $n$
and $\vec{x}$ is an n-tuple of pairwise distinct variables,
 and $\phi$ is an \La-constraint ranging over the variables mentioned.
\end{de}

Constraint languages have to be closed under variable renaming, closed under
intersection, and the satisfiability problem of such languages has to be decidable.

A \textbf{goal} is defined as a possibly empty conjunction of
\La-constraints and \Rl-atoms. Relying on conventional logical
terminology, a \textbf{\Po-answer} of a goal $G$ for a
program \Po  can be defined as a satisfiable \La-constraint $\phi$
s.t.\ the implication $\phi \rightarrow G$ is a logical consequence of \Po.

SLD-resolution is generalized by performing goal reduction only on the
\Rl-atoms and solving conjunctions of collected \La-constraints by
the \La-constraint solver.
\textbf{Goal reduction} is managed by a binary relation \gr on the set
of goals as follows (\Xsf{V} denotes the finite set of variables in
the query and $\Xsf{V}(\cdot)$ is a function assigning to a constraint
the finite set of variables constrained by it).

{\em
\begin{center}
$A \: \& \: G \gr F \: \& \: G$ 
if $A \leftarrow F$ is a variant of a clause in \Po \\
s.t. $(\Xsf{V} \cup \Xsf{V}(G)) \cap \Xsf{V}(F) \subseteq \Xsf{V}(A)$.
\end{center}}

A second rule takes care of constraint solving for the \La-constraints
appearing in subsequent goals. The rule takes the conjunction of the
\La-constraints from the reduced goal and the applied clause and
gives, via the black box of a suitable \La-constraint solver, a
satisfiable \La-constraint in solved form if the conjunction of
\La-constraints is satisfiable.  The \textbf{constraint solving} rule
can then be defined as a total function \cs on the set of goals as
follows ($CS(\cdot)$ denotes the \La-constraint solver as a function
on the set of \La-constraints).

{\em
\begin{center}
$\phi \:\&\: \phi' \:\&\: G \cs \phi'' \:\&\: G$ 
if $CS(\phi \:\&\: \phi') = CS(\phi'')$.
\end{center}}

\citeN{HuS:88} show that this generalized SLD-resolution method is a
sound and complete method for inferring \Po-answers.
For the following discussion it will be convenient to view this
operational semantics as a search of a tree.
For a given query and a given program, the search space determined by
the derivation rules \gr and \cs can be described as a derivation tree as follows.

\begin{de}[derivation tree] A derivation tree determined by a query
  $G_1$ and a definite clause specification \Po has to satisfy the
  following conditions:
\begin{enumerate}
\item Each node is either a relation-node or a constraint-node.
\item The descendants of every relation-node are all constraint-nodes
  s.t.\ for every \gr-resolvent $G'$ obtainable by a clause $C$
  from goal $G$ in a relation-node, there is a descending constraint-node labeled
  by $C$ and $G'$.
\item The descendants of every constraint-node are all relation-nodes
  s.t.\ for every unique \cs-resolvent $G \:\&\: \phi''$ obtainable
  from goal $G \:\&\: \phi \:\&\: \phi'$ in a constraint-node, there
  is a descending relation-node labeled by $G \:\&\: \phi''$.
\item The root node is a relation-node labeled by $G_1$.
\item A success node is a terminal relation-node labeled by a
  satisfiable \La-constraint.
\end{enumerate}
\end{de}

Successful derivations correspond to certain subtrees of derivation
trees and can be defined as proof trees as follows.

\begin{de}[proof tree] A proof tree for a query $G_1$ from \Po is a subtree of
  a derivation tree determined by $G_1$ and \Po and is defined as
  follows:
\begin{enumerate}
\item A relation-node of the proof tree is a relation-node of the supertree and
  takes \emph{one} of the descendants of the supertree relation-node as
  its descendant.
\item A constraint-node of the proof tree is a constraint-node of the
  supertree and takes the unique descendant of the supertree constraint-node as its
  descendant.
\item The root node of the proof tree is the root node of the
  supertree.
\item The terminal node of the proof tree is a success node of the
  supertree labeled by a satisfiable \La-constraint, called answer constraint.
\end{enumerate}
\end{de}

\section{Baum's Maximization Technique and Probabilistic CLP} \label{Baum}

One straightforward way to add statistical information to symbol
processing models is to define the derivation process of such models as
a a stochastic process as follows: Make a
stochastic choice at each derivation step and assume the stochastic
choices to be independent of each other. Calculate the
probability of a derivation as the joint probability of the
independent stochastic choices made and the probability of an input 
as the sum of the probabilities of its derivations.
This is the probabilistic model underlying, e.g., Hidden Markov Models or
stochastic context-free grammars. The parameters of such models, i.e.,
the probabilities of the stochastic choices, can be estimated by
Baum's maximization technique \cite{Baum:67,Baum:70,Baum:72}.

The basic formal concepts of this
technique can be described in an abstract way as follows:
Let $\Pi = \{ \pi_{ij} \}$
be the parameter set of an abstract probabilistic symbol processing model where
$\pi_{ij} \geq 0$ and $\sum_j \pi_{ij} = 1$.
The variable $i$ ranges over the types of choices that the stochastic process
makes and the variable $j$ ranges over the
alternatives to choose from when a choice of type $i$ is made.
Furthermore, let $y$ denote an input of the
probabilistic processing model, i.e., an observation sequence, and let
$x$ denote an output of the model, i.e., an analysis, and let $Y(x) = y$ be the unique observation
corresponding to analysis $x$ and $X(y) = \{ x | Y(x) = y \}$ be the
set of analyses of observation $y$. Finally, let $\nu_{ij}(x)$ be the
number of selections of alternative $j$ for a choice of type $i$ in analysis
$x$.
Then the probability of an analysis can be calculated as the product
of the probabilities of the stochastic choices made in producing it:

\begin{center}
$p(x;\pi) = \prod_{ij} \pi_{ij}^{\nu_{ij}(x)}$
\end{center}
The probability of an observation then is the sum of the probabilities
of its analyses:

\begin{center}
$p(y;\pi) = \sum_{x \in X(y)} p(x;\pi)$
\end{center}
The purpose of Baum's maximization technique is to find maximum
likelihood parameter values, i.e., $\{ \pi_{ij} \}$ which maximize
the likelihood function $P(\pi) = \prod_y p(y;\pi)$ for a
given $y$-sample. To this end Baum defines a transformation $\tau$ of
$\pi$ into itself, which looks in its basic form as follows:

\begin{center}
$\tau(\pi_{ij}) =
\frac{\sum_y N_{ij}}{\sum_y \left( \sum_k N_{ik} \right)} =
\frac{\sum_y \left( \sum_x p(x | y) \nu_{ij}(x) \right)}
{\sum_y \left( \sum_k \left( \sum_x p(x | y) \nu_{ik}(x)
\right) \right) }$,
\end{center}
and $\tau$ yields an iterative algorithm where each step is defined
by

\begin{center}
$\pi_{ij}^{t+1} = \tau(\pi_{ij}^{t})$.
\end{center}
This algorithm is hill-climbing, i.e., it can be shown that 
$P(\tau(\pi)) > P(\pi)$
unless $\tau(\pi)$ is a critical
point of $P$ or equivalently is a fixed point of $\tau$.

Attempts to apply this algorithm and the underlying abstract model
directly to a model of probabilistic CLGs or CLP were presented,
e.g., by \citeN{Eisele:94}, \citeN{Miyata:96} and
\citeN{Osborne:97}. A detailed critique of such attempts with respect
to the problem of parameter estimation from complete data can be found
in \citeN{AbneySAVG:96}. What Abney calls the Expected Rule Frequency
(ERF) parameter estimation method can be seen as a special case of
Baum's maximization technique. In order to show that Baum's general
algorithm fails as an optimization technique for the maximum
likelihood problem for probabilisitic CLP, we simply can give a
counterexample using a deterministic program. In this case, parameter
estimation from incomplete data using Baum's method is equivalent to
using the ERF method. This point shall be made explicit in the following.

Let us apply the above-defined abstract model to a simple
deterministic constraint logic program. The stochastic choices of the
abstract model correspond to application probabilities of definite
clauses in the generalized SLD-resolution procedure;
the alternatives to choose from
when an atom is selected in goal reduction are the different clauses
defining the selected atom. In the following example (see
Fig. \ref{program}), each clause will be annotated by a
choice-alternative pair indicating a probabilisitic parameter $\pi_{ij}$.

\begin{figure}[htbp]
\begin{center}
\begin{tabular}{l}
\texttt{11} $s(Z) \leftarrow p(Z) \:\&\: q(Z).$ \\
\texttt{21} $p(Z) \leftarrow Z=a.$ \\
\texttt{22} $p(Z) \leftarrow Z=b.$ \\
\texttt{31} $q(Z) \leftarrow Z=a.$ \\
\texttt{32} $q(Z) \leftarrow Z=b.$ 
\end{tabular}
\end{center}
\caption{A sample program}
\label{program}
\end{figure}

\begin{figure}[htbp]
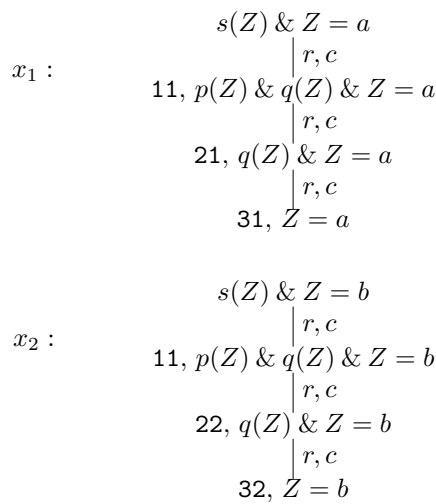

\begin{center} 
\begin{tabular}{p{75pt}c}

$x_1 :$
 
&

\begin{bundle}{$s(Z) \:\&\: Z=a$}
\chunk[$\qquad r,c$]
{
\begin{bundle}{\texttt{11}, $p(Z) \:\&\: q(Z) \:\&\: Z=a$}
\chunk[$\qquad r,c$]
{
\begin{bundle}{\texttt{21}, $q(Z) \:\&\: Z=a$}
\chunk[$\qquad r,c$]
{\texttt{31}, $Z=a$}
\end{bundle}
}
\end{bundle}
}
\end{bundle}

\end{tabular}

\end{center}

\vspace{3ex}

\begin{center} 
\begin{tabular}{p{75pt}c}

$x_2 : $
 
&

\begin{bundle}{$s(Z) \:\&\: Z=b$}
\chunk[$\qquad r,c$]
{
\begin{bundle}{\texttt{11}, $p(Z) \:\&\: q(Z) \:\&\: Z=b$}
\chunk[$\qquad r,c$]
{
\begin{bundle}{\texttt{22}, $q(Z) \:\&\: Z=b$}
\chunk[$\qquad r,c$]
{\texttt{32}, $Z=b$}
\end{bundle}
}
\end{bundle}
}
\end{bundle}

\end{tabular}

\end{center}

\caption{Proof trees from sample program}
\label{prooftrees}
\end{figure}

The relational atom $s(Z)$ is defined uniquely in clause \texttt{11}.
The atoms $p(Z)$ and $q(Z)$ each are defined in two different ways,
which for the sake of the example are considered to be incompatible.
For a selection of atom $p(Z)$ one can choose between clauses
\texttt{21} and \texttt{22} in a goal reduction step,
whereas  for a choice of atom $q(Z)$ the alternatives to choose from are clauses
\texttt{31} and \texttt{32}.
This program is deterministic for the queries $s(Z) \:\&\:
Z=a$ and $s(Z) \:\&\: Z=b$. This means, there is only one proof tree
from the above program for each query (see Fig. \ref{prooftrees}).
The proof tree $x_1$ for the query $s(Z) \:\&\: Z=a$ uses clauses
\texttt{11, 21} and \texttt{31} and yields answer constraint $Z=a$;
the proof tree $x_2$ for the second query uses clauses \texttt{11, 22} and
\texttt{32} and gives answer constraint $Z=b$.

Let us now consider the application of Baum's maximization technique
to estimate the parameters of such a probabilistic CLP model (see
Fig. \ref{estimation}). An input corpus consisting of the three queries $y_1:
s(Z) \:\&\: Z=a, \; y_2: s(Z) \:\&\: Z=a$ and $ y_3: s(Z) \:\&\: Z=b$
will yield the corresponding unique proof trees $x_1 \in X(y_1), \;
x_1 \in X(y_2)$ and $x_2 \in X(y_3)$. 
The conditional probabilities $p(x|y)$ for $x \in X(y)$ will be 1 in
each case since there is a unique proof tree for each query.
Thus for the calculation of
$N_{ij} = \sum_x p(x | y) \nu_{ij}(x)$,
the expected number of occurences of clauses in proof trees, we simply
have to count and can ignore the respective probabilities of the proof
trees. The algorithm then will give unique estimated parameter values
$\hat \pi_{ij} = \frac{\sum_y N_{ij}}{\sum_y  \left( \sum_k N_{ik}
\right)}$
immediately.

\begin{figure}
\begin{center}
\begin{tabular}{c c c|c|c c|c c|}
$y$ &
$x \in X(y)$ &
$p(x|y)$  &
$N_{11}$ &
$N_{21}$ &
$N_{22}$ &
$N_{31}$ &
$N_{32}$ \\
\hline
$ y_1$ & $x_1$ & 1 & $1 \cdot1$ & $1 \cdot1$ & $1 \cdot0$  & $1 \cdot1$ &  $1 \cdot 0$ \\
$y_2$ & $x_1$ & 1 & $1 \cdot1$ & $1 \cdot1$ & $1 \cdot0$  & $1 \cdot1$ & $1 \cdot0$  \\
$y_3$ & $x_2$ & 1 & $1 \cdot1$ & $1 \cdot0$ & $1 \cdot1$ & $1 \cdot0$  & $1 \cdot1$ \\
\hline
 & & $\sum_y N_{ij}=$ & 3 & 2 & 1 & 2 & 1 \\
 & & $\sum_y \sum_k N_{ik}=$ & 3 & 3 & 3 & 3 & 3\\
\hline
 & & $\hat\pi_{ij}=$ & 1 & 2/3 & 1/3 & 2/3 & 1/3
\end{tabular}
\end{center}
\caption{A sample estimation}
\label{estimation}
\end{figure}

If we now consider the calculation of the probability distribution
over the proof trees of such a probabilistic CLP model, we see that in
contrast to the above-defined abstract model we cannot simply calculate a
product for each proof tree. Instead, in order to get a proper
probability distribution over proof trees, we have to do an additional
normalization. For example, if the sum of the unnormalized
probabilities of the proof trees under the estimated model, 
$p(x_1; \hat\pi) + p(x_2; \hat\pi) = 4/9 + 1/9 = 5/9$,
is used as a normalization constant, then we will get a normalized
probability distribution over proof trees, 
$p'(x_1;\pi') = 4/5, \; p'(x_2;\pi') = 1/5$,
yielding a normalized likelihood of our training corpus
$P'(\pi') = (4/5)^2 \cdot 1/5 = .128$. 
Note that the normalized probability distribution no longer refers to
specific parameter values. In fact, there is no analytical solution to the
problem of finding parameter values $\pi'$ for the program of
Fig. \ref{program} which yield probability distribution
$p'$ over the proof trees of Fig. \ref{prooftrees}.
However, given the same preconditions, we can find
a probability distribution 
$p''(x_1;\pi'') = 2/3, \; p''(x_2; \pi'') = 1/3$ which
yields a higher likelihood
$P''(\pi'') = (2/3)^2 \cdot1/3 = .148$.
This contradicts the assumption that the parameter values estimated
by Baum's technique are the requested maximum likelihood
estimates for a probabilistic CLP model as defined above.

\section{A Log-linear Model for Probabilistic CLP}
\label{LogLinearModel}

The above-discussed approach based a probability distribution over
proof trees on a definition of the derivation process of CLP as a
(context-free) stochastic process. An alternative, presented by
\citeN{AbneySAVG:96}, for his model of
stochastic attribute-value grammars is to define a probability
distribution over dags as a random field. This probability model does
not build on any underlying stochastic process but rather on the
underlying graphical structure of the analyses produced by the model.
Random fields can be seen as special instances of general log-linear
probability models. Such a model can be defined as follows.

\begin{de}[log-linear distribution]\label{LogLinearDistribution}
A log-linear probability distribution $p_{\lambda \cdot \nu}$ on a set
  $\Omega$ is defined s.t.\ for all $\omega \in \Omega$:
\begin{center}
$p_{\lambda \cdot \nu}(\omega) =
{Z_{\lambda \cdot \nu}}^{-1}
e^{\lambda \cdot \nu(\omega)} p_0(\omega)$
\end{center}
where 
$Z_{\lambda \cdot \nu} = \sum_{\omega \in \Omega} 
e^{\lambda \cdot \nu(\omega)} p_0(\omega)$
is a normalizing constant,\\
$\lambda = (\lambda_1, \ldots, \lambda_n)$
is a vector of log-parameters s.t.\ $\lambda \in \Rn$,\\
$\mathbf{\chi} = (\chi_1, \ldots, \chi_n)$
is a vector of properties, \\
$\mathbf{\nu} = (\nu_1, \ldots, \nu_n)$
is a vector of property-functions s.t. for each \\
$\nu_i:\Omega
\rightarrow \Nats $, $\nu_i(\omega)$
is the number of occurences of property $\chi_i$ in $\omega$, \\
$\lambda \cdot \nu (\omega)$
is a weighted property-function s.t. 
$\lambda \cdot \nu (\omega) =
\sum^n_{i=1}\lambda_i \nu_i(\omega)$,\\
$p_0$ is a fixed initial distribution.
\end{de}
In analogy to stochastic attribute-values grammars, we can define a
probability distribution over proof trees as a special log-linear
model. The special instance of interest is simply a log-linear
distribution on the countably infinite set of proof-trees for a set of
queries to a program. Such a distribution is determined by a vector of
properties and a corresponding vector of log-parameters. Properties
could be defined,  e.g., as subtrees of proof trees. For the moment,
we can leave an exact definition of properties aside and refer to an
assumed vector of property-functions.

The form of log-linear models can be rationalized as an example
of an \textbf{exponential family} of probability functions. From this
viewpoint this model can be seen as just a very flexible probability
model defining the probability of a configuration to be proportional
to the product of weights assigned to arbitrary properties of the configuration.

\begin{center}
$p(\omega) \propto \prod_{i=1}^n \pi_i^{\nu_i(\omega)}$.
\end{center}
This can be put in the form of Definition
\ref{LogLinearDistribution} by replacing proportionality by a constant
and parameters $\pi_i$ by log-parameters
$\lambda_i=log \: \pi_i$.

\begin{eqnarray*}
p(\omega) & = & C \prod\nolimits_{i=1}^n \pi_i^{\nu_i(\omega)} \\
 & = & Z^{-1} \prod\nolimits_{i=1}^n \pi_i^{\nu_i(\omega)} \\
 & = & Z^{-1} \prod\nolimits_{i=1}^n e^{\lambda_i \nu_i(\omega)} \\
 & = & Z^{-1} e^{\sum_{i=1}^n \lambda_i \nu_i(\omega)} .
\end{eqnarray*}

Another way to rationalize the form of the log-linear model is as a
\textbf{maximum entropy} probability distribution. From this
viewpoint we do statistical inference and, believing that entropy is
the unique consistent measure of the amount of uncertainty represented
by a probability distribution, we obey the following principle: 
\begin{quote}
In making inferences on the basis of partial information we must use that
probability distribution which has maximum entropy subject to
whatever is known. This is the only unbiased assignment we can make;
to use any other would amount to arbitrary assumption of information
which by hypothesis we do not have. \cite{Jaynes:57}
\end{quote}
More formally, suppose a random variable $X$ can take on values $x_i,
i=1, \ldots, n$ and we want to estimate the corresponding
probabilities $p_i, i=1, \ldots, n$. All we have are expectations of
functions $f_k(X), k=1,\dots, m$.
The maximum entropy principle can then be stated as follows.

{\em
\begin{quote}
Maximize $H(p_1, \ldots, p_n) =
- \sum_{i=1}^n p_i \: log \: p_i$ subject to the constraints
$\sum_{i=1}^n p_i f_k(x_i) = F_k, k = 1, \ldots, m$ and $\sum_{i=1}^n
p_i = 1$.
\end{quote}}
The solution we get for all $p_i, i=1, \ldots, n$ is:

\begin{center}
$p_i = \frac{\textstyle e^{\sum_{k=1}^m \lambda_k f_k(x_i)}}
{\textstyle \sum_{i=1}^n e^{\sum_{k=1}^m \lambda_k f_k(x_i)}}$.
\end{center}
This result follows directly from a constrained optimization argument
where the parameters are viewed as Lagrange multipliers:

Let $\Lambda$ denote the Lagrangian defined by
$\Lambda(p_1, \ldots, p_n, \lambda_0, \lambda_1, \ldots, \lambda_m) =
-\sum_{i=1}^n (p_i \: log \: p_i )+
(\lambda_0 +1) \sum_{i=1}^n (p_i -1) +
\lambda_1 \sum_{i=1}^n (p_i f_1(x_i) - F_1) +
\cdots +
\lambda_m \sum_{i=1}^n (p_i f_m(x_i) - F_m).$ \\
Then $\frac{\partial}{\partial p_i} \Lambda = 
- (log \: p_i + 1) + (\lambda_0 +1) + \lambda_1 f_1(x_i) + \cdots +
\lambda_m f_m(x_i)$. \\
Set $\frac{\partial}{\partial p_i} \Lambda = 0$, then
$p_i = e^{\lambda_0 + \sum_{k=1}^m \lambda_k f_k(x_i)}$. \\
Since $\sum_{i=1}^n p_i =1$, we have 
$e^{\lambda_0} \sum_{i=1}^n e^{\sum_{k=1}^m \lambda_k f_k(x_i)} =
1$. \\
Define $Z = \sum_{i=1}^n e^{\sum_{k=1}^m \lambda_k f_k(x_i)}$, 
then $\lambda_0 = log \: Z^{-1}$ \\
and $p_i = Z^{-1} e^{\sum_{k=1}^m \lambda_k f_k(x_i)} = 
\frac{e^{\sum_{k=1}^m \lambda_k f_k(x_i)}}
{\sum_{i=1}^n e^{\sum_{k=1}^m \lambda_k f_k(x_i)}}$.

Log-linear models originated in statistical physics as flexible
probabilistic models of equilibrium states of physical systems. Jaynes
interpreted such Gibbs- or Boltzmann-distributions in a more abstract
maximum-entropy framework (see \citeN{Jaynes:83}). 
Besides numerous applications in the area of natural language
processing\footnote{The applications include, beside others, 
  probabilistic grammar models \cite{Mark:92,AbneySAVG:96},
word morphology \cite{Della-Pietra:95},
machine translation \cite{Berger:96},
language modelling \cite{Rosenfeld:96},
part-of-speech tagging \cite{Ratnaparkhi:96},
word correlations \cite{LaffertyACL:97}
and text segmentation \cite{LaffertyEMNLP:97}.},
log-linear models are also applied successfully in image processing (see
the work on random fields initiated by \citeN{Geman:84})
and are closely related to other probabilistic
models such as Boltzmann machines (see \citeN{Ackley:85}) or graphical
models  (see \citeN{Pearl:88}).

The work presented in the following sections applies for the most part
to log-linear models in general. We will refer for this discussion to
Definition \ref{LogLinearDistribution}. In case the property vector is
fixed and clear from the context, the model will be written
$p_\lambda$ to indicate the depencence on the parameter vector.
Furthermore, it will be convenient to have a recursive definition of models based on
property-functions which are extended by additional properties and corresponding
parameters or by new parameters.

\begin{po} For each weighted property-funtion
$\phi(\omega) = \mathbf{\lambda \cdot \nu}(\omega)$, \\
$\psi(\omega) = \mathbf{\gamma \cdot \mu}(\omega)$
(with possibly $\mathbf{\nu} = \mathbf{\mu}$),
let $(\psi + \phi)(\omega) = \psi(\omega) + \phi(\omega)$ be an extended
property-function (reducing to $\mathbf{\lambda} +\mathbf{\gamma}$ in
case $\mathbf{\nu} = \mathbf{\mu}$).
Then 
$p_{\psi + \phi}(\omega) = {Z_{\psi + \phi}}^{-1}
e^{\psi(\omega)} p_{\phi}(\omega)$ 
where $Z_{\psi + \phi} =
\sum_{\omega \in \Omega} e^{\psi(\omega)} p_{\phi}(\omega)$.
\end{po}

\begin{proof}
\begin{eqnarray*}
p_{\psi + \phi}(\omega)
& = & 
{Z_{\psi + \phi}}^{-1} e^{\psi + \phi(\omega)}
p_0(\omega) \\
& = & 
(\sum_{\omega \in \Omega} e^{\psi(\omega) + \phi(\omega)}
p_0(\omega) )^{-1} e^{\psi(\omega) + \phi(\omega)} p_0(\omega)\\
& = & 
( \sum_{\omega \in \Omega} e^{\psi(\omega)} e^{\phi(\omega)}
p_0(\omega) Z_{\phi} {Z_{\phi}}^{-1} ) ^{-1} e^{\psi(\omega)}
e^{\phi(\omega)} p_0(\omega)\\
& = & 
{Z_\phi}^{-1} (\sum_{\omega \in \Omega} e^{\psi(\omega)}
p_{\phi}(\omega) )^{-1}
e^{\psi(\omega)} e^{\phi(\omega)} p_0(\omega) \\
& = & 
(\sum_{\omega \in \Omega} e^{\psi(\omega)}  p_{\phi}(\omega)
)^{-1} e^{\psi(\omega)} p_{\phi}(\omega).
\qed
\end{eqnarray*}
\renewcommand{\qed}{}
\end{proof}

\section{Inducing Log-linear Models from Incomplete
  Data}\label{Induction}

Induction of log-linear models involves two problems: parameter
estimation and property selection. In the following we will give a detailed
presentation of solutions to these problems for the case of incomplete
data.

\subsection{Parameter Estimation from Incomplete Data}
\label{ParameterEstimation}

An algorithm to estimate the parameters of general log-linear models
from complete data has been presented by
\citeN{Della-Pietra:95}. Their ``Improved Iterative Scaling''
algorithm is an extension of the ``Generalized Iterative Scaling''
algorithm of \citeN{Darroch:72} especially tailored to estimating
models with large parameter spaces. The algorithm is a technique for
maximum likelihood estimation for log-linear models from complete
data, i.e., it addresses the problem of  maximizing the complete-data
log-likelihood function $log \: \prod_x p(x)^{\tilde{p}(x)}$ for a
given empirical distribution $\tilde{p}(x)$ over complete data $x$. The
solution to this problem is equivalent to the solution to the maximum
entropy problem subject to linear constraints, i.e., the problem of
maximizing the entropy $H(p)$ subject to the constraints $\sum_x p(x)
f_k(x) = \sum_x \tilde{p}(x) f_k(x), \; k = 1, \ldots, m$ with respect
to the complete data empirical expectation (see
\citeN{Della-Pietra:95}). In the language of constrained optimization,
the maximum likelihood problem for log-linear models with respect to complete
data is the dual to the maximum entropy problem for linear constraints
with respect to complete data (see \citeN{Berger:96}).

However, the need to rely on large training samples of complete
data may be inconvenient if complete data are complex and
difficult to gather. This is the case for applications of CLP to
natural language processing. Here complete data means several
person-years of hand-annotating large corpora with detailed analyses
of specialized grammar frameworks. Clearly, for such applications
parameter estimation from incomplete data, i.e., unanalyzed input of
natural language strings, is desirable.

Unfortunately, Iterative Scaling will no longer
work if the training data are incomplete. The
incomplete-data log-likelihood takes the form 
$log \prod_y \sum_{x \in X(y)} p(x)$, i.e., the probability the model
assigns to the data strings is the product of the probabilities of the
strings and the probability of a string  is calculated as the sum of the
probabilities of its analyses. In contrast to the complete-data
log-likelihood this function is non-concave (it involves a sum inside
the logarithm) and cannot be maximized directly or uniquely.

In the following we will show how the numerical algorithm of
\citeN{Della-Pietra:95} can be redefined in order to fit incomplete
data. The new algorithm can be defined in the EM-framework of maximum
likelihood estimation from incomplete data of \cite{Dempster:77}.
Applying this framework to the problem of probabilistic CLP, we can
assume the following to be given:

{\em
\begin{itemize}
\item Observed, incomplete data $y \in \mathcal{Y}$
corresponding to a given, finite set of queries for a
constraint logic program \Po, 
\item Unobserved, complete data $x \in \mathcal{X}$ corresponding to
  the countably infinite set of proof trees for queries $\mathcal{Y}$ from a
  constraint logic program \Po,
\item Functions $Y:\mathcal{X} \rightarrow \mathcal{Y}$
s.t. $Y(x) = y$ corresponds to the unique query labeling proof tree $x$, and
$X:\mathcal{Y} \rightarrow \mathcal{X}$
s.t. $X(y) = \{x |\; Y(x) = y \}$
is the countably infinite set of proof trees for query $y$ from a constraint logic
program \Po,
\item Complete data specifications $p_{\lambda}$ 
s.t. $p_\lambda (x)$ is a log-linear distribution on $\mathcal{X}$
with given initial distribution $p_0$,
fixed properties $\chi$ and
property-functions vector $\nu$
and depending on parameter vector $\lambda$, 
\item Incomplete data specifications $L$ s.t.
$L(\lambda)
= log \prod_{y \in \mathcal{Y}} \sum_{x \in X(y)} p_{\lambda}(x) $ 
$= \sum_{y \in \mathcal{Y}} log \sum_{x \in X(y)} p_{\lambda}(x) $
$= \sum_{y \in \mathcal{Y}} log \; p_{\lambda}(y)$
is the log-likelihood of a fixed $\mathcal{Y}$-sample depending on
parameter vector $\lambda$.
\end{itemize}}

For the discussion of parameter estimation we will refer to a given
vector of property functions. This is assumed to
result from the property selection procedure defined in
Sect. \ref{PropertySelection}, whereby for each property function
$\nu_i$ some proof tree $x \in \mathcal{X}$ s.t. $\nu_i(x) > 0$
is assumed to exist. Furthermore, we require $p_\lambda$ to be
strictly positive on $\mathcal{X}$, i.e., $p_\lambda(x) > 0$ for all $x
\in \mathcal{X}$.

The problem of maximum likelihood estimation of log-linear models from
incomplete data can then be stated formally as follows.

{\em
\begin{quote}
Given a fixed $\mathcal{Y}$-sample and a set $\Lambda =
\{ \lambda |\; p_{\lambda}(x)$ is a log-linear distribution on
$\mathcal{X}$ with fixed $p_0$, fixed $\nu$ and
$\lambda \in \Rn \}$,
we want to find the maximum likelihood estimate
$\lambda^\ast \in \Lambda$  s.t. 
$\lambda^\ast = argmax_{\lambda \in \Lambda} L(\lambda)$.
\end{quote}}

The key idea of the following approach is to iteratively maximize a
strictly concave auxiliary function when the log-likelihood objective
function cannot be maximized analytically. An auxiliary function
convenient for our problem can be defined as a two-place function $A$
giving an estimate of the improvement in the incomplete-data log-likelihood $L$
when going from a model $p_\lambda$ to a model
$p_{\gamma + \lambda}$.

In the following
$p[f] = \sum_{\omega \in \Omega} p(\omega) f(\omega)$
will denote the expectation of a function $f:\Omega \rightarrow
\Reals$ with respect to a probability distribution
$p$ on a set $\Omega$.

\begin{de} Let $\lambda \in \Lambda$,
  $\gamma \in \Rn$.
Then 
\begin{center}
$A(\gamma + \lambda) = \sum_{y \in \mathcal{Y}} 
( 1 + k_{\lambda} \left[ \gamma \cdot \nu \right] -
p_{\lambda} \left[ \sum^n_{i=1} {\bar{\nu}}_i e^{\gamma_i \nu_{\#} }\right] 
)$ 
\end{center}
where ${\bar{\nu}}_i(x) = \frac{\nu_i(x)}{\nu_{\#}(x)}$,
$\nu_{\#}(x) = \sum^n_{i=1} \nu_i(x)$,
$k_{\lambda}(x) = \frac{p_{\lambda}(x)}{\sum_{x \in X(y)}
 p_{\lambda}(x)}$.
\end{de}
By considering the first and second derivative of $A$, we see that
$A$ is strictly concave in the parameters. Strict concavity together
with continuity of the function and closedness of the parameter space
directly gives us a unique maximum of $A$.

\begin{po} For each $\lambda \in \Lambda$, $\gamma \in \Rn$:
$A(\gamma + \lambda)$ takes its maximum as a function of
  $\gamma$ at the unique point $\hat\gamma$ satisfying for each
  $\hat\gamma_i, i=1, \ldots, n$:
\begin{center}
$\sum_{y \in \mathcal{Y}}  k_{\lambda} \left[ \nu_i \right] 
= \sum_{y \in \mathcal{Y}} p_{\lambda} \left[ \nu_i
    e^{\hat\gamma_i \nu_{\#} }\right] .$
\end{center}
\end{po}

\begin{proof}
\begin{eqnarray*}
\frac{\partial}{\partial \gamma_i}
A(\gamma + \lambda)
& = & 
\frac{\partial}{\partial \gamma_i}
\sum_{y \in \mathcal{Y}} 
( 1 + k_{\lambda} [ \gamma \cdot \nu ] -
p_{\lambda} [ \sum^n_{j=1} {\bar{\nu}}_j e^{\gamma_j \nu_{\#} }] 
) \\
& = & 
\sum_{y \in \mathcal{Y}} 
( \frac{\partial}{\partial \gamma_i}
\sum^n_{j=1}
( \frac{1}{n} + k_{\lambda} [ \gamma_j \cdot \nu_j ] -
p_{\lambda} [  {\bar{\nu}}_j e^{\gamma_j \nu_{\#} }] 
) )\\
& = &  
\sum_{y \in \mathcal{Y}} 
( \sum_{j \not= i} 
( \frac{\partial}{\partial \gamma_i}
( \frac{1}{n} + k_{\lambda} [ \gamma_j \cdot \nu_j ] -
p_{\lambda} [  {\bar{\nu}}_j e^{\gamma_j \nu_{\#} }] 
) ) \\
& & 
+ \frac{\partial}{\partial \gamma_i}
( \frac{1}{n} + k_{\lambda} [ \gamma_i \cdot \nu_i ] -
p_{\lambda} [  {\bar{\nu}}_i e^{\gamma_i \nu_{\#} }] 
) ) \\
& = & 
\sum_{y \in \mathcal{Y}} 
( k_{\lambda} [ \nu_i ] 
- \sum_{x \in \mathcal{X}} 
(  p_{\lambda} (x) \bar\nu_i (x) e^{\gamma_i \nu_{\#}(x) } \nu_{\#}(x)
) ) \\
& = & 
\sum_{y \in \mathcal{Y}} 
( k_{\lambda} [ \nu_i ] 
- \sum_{x \in \mathcal{X}} 
( p_{\lambda} (x) \nu_i (x) e^{\gamma_i \nu_{\#}(x)}
) ) \\
& = & 
\sum_{y \in \mathcal{Y}} 
( k_{\lambda} [ \nu_i ] 
- p_{\lambda} [ \nu_i e^{\gamma_i \nu_{\#}} ] ).\\
& & \\
\frac{\partial^2}{{\partial \gamma_i}^2} A(\gamma + \lambda)
& = & 
\frac{\partial}{\partial \gamma_i}
\sum_{y \in \mathcal{Y}} 
( k_{\lambda} [ \nu_i ] 
- p_{\lambda} [ \nu_i e^{\gamma_i \nu_{\#}} ] ) \\
& = & 
- \sum_{y \in \mathcal{Y}} 
(
\frac{\partial}{\partial \gamma_i} p_{\lambda} [ \nu_i e^{\gamma_i \nu_{\#}}
] 
)\\
& = & 
- \sum_{y \in \mathcal{Y}} 
( \sum_{x \in \mathcal{X}} 
( p_{\lambda}(x) \nu_i(x) e^{\gamma_i
    \nu_{\#}(x)} \nu_{\#}(x) ) )\\
& = & 
- \sum_{y \in \mathcal{Y}} 
p_{\lambda} [ \nu_i \nu_\# e^{\gamma_i \nu_\#} ]  \\
& < & 0.
\qed
\end{eqnarray*}
\renewcommand{\qed}{}
\end{proof}

At the core of the proposed method lies the definition of an iterative
algorithm for maximizing $L$ which is constructed from the auxiliary
function $A$. At each step of this ``Iterative Maximization (IM)'' algorithm
a model based on parameters $\lambda$ is extended by a parameter
vector $\hat\gamma$ which gives the maximum estimated improvement in
log-likelihood $L$, i.e., which is obtained by maximizing the
auxiliary function $A(\gamma + \lambda)$ as a function of $\gamma$.

\begin{de}[iterative maximization] Let $\mathcal{M}:\Lambda
  \rightarrow \Lambda$ be a mapping defined by
\begin{quote}
$\mathcal{M}(\lambda) = \hat\lambda \in \Lambda$ s.t.
$\hat \lambda = \hat \gamma + \lambda$ with 
$\hat \gamma = argmax_{\gamma \in \Rn} A(\gamma + \lambda)$.
\end{quote}
Then each step of the Iterative Maximization Algorithm is defined by
\begin{quote}
$\lambda^{(k+1)} = \mathcal{M}(\lambda^{(k)})$.
\end{quote}
\end{de}

To show the central convergence properties of the IM algorithm, we
first have to show some provisional results.
Lemma \ref{A<L-L} shows that the auxiliary function
$A(\gamma + \lambda) $ is a lower bound on 
$L(\gamma + \lambda) - L(\lambda)$, 
the difference in log-likelihood between the basic
and the extended model, i.e., it is a conservative estimate of the
improvement in log-likelihood.

\newtheorem{lemma}[po]{Lemma}
\begin{lemma} \label{A<L-L}
$A(\gamma + \lambda) \leq L(\gamma + \lambda) - L(\lambda)$.
\end{lemma}

\begin{proof}
\begin{eqnarray*}
L(\gamma + \lambda) - L(\lambda)
& = &
\sum_{y \in \mathcal{Y}} (
log \frac{p_{\gamma +\lambda}(y)}{p_{\lambda}(y)}
) \\
& = & 
\sum_{y \in \mathcal{Y}} (
log \frac{1}{p_{\lambda}(y)} 
\sum_{x \in X(y)} (
p_{\gamma + \lambda}(x) 
\frac{p_{\lambda}(x)}{p_{\lambda}(x)}
)
) \\
& = & 
\sum_{y \in \mathcal{Y}} (
log \sum_{x \in X(y)} (
\frac{p_{\lambda}(x)}{p_{\lambda}(y)}
\frac{p_{\gamma + \lambda}(x)}{p_{\lambda}(x)}
)
) \\
& \geq & 
\sum_{y \in \mathcal{Y}} (
\sum_{x \in X(y)} (
\frac{p_{\lambda}(x)}{p_{\lambda}(y)}
log \frac{p_{\gamma + \lambda}(x)}{p_{\lambda}(x)}
)
) \\
& & \textrm{ by Jensen's inequality} \\
& = & 
\sum_{y \in \mathcal{Y}} (
\sum_{x \in X(y)} (
\frac{p_{\lambda}(x)}{p_{\lambda}(y)} (
log \; p_{\gamma + \lambda}(x) 
- log \; p_{\lambda}(x)
)
)
)\\
& = & 
\sum_{y \in \mathcal{Y}} (
\sum_{x \in X(y)} (
\frac{p_{\lambda}(x)}{p_{\lambda}(y)} (
log \; Z_{\gamma + \lambda}^{-1}
+ log \; e^{\gamma \cdot \nu(x)} 
\\
& & 
+ log \;  p_{\lambda}(x)
- log \;  p_{\lambda}(x)
))) \\
& = & 
\sum_{y \in \mathcal{Y}} (
k_{\lambda} [ \gamma \cdot \nu ]
- log \; p_{\lambda} [ e^{\gamma \cdot \nu} ]
)\\
& \geq & 
\sum_{y \in \mathcal{Y}} (
k_{\lambda} [ \gamma \cdot \nu ]
+1 
-  p_{\lambda} [ e^{\gamma \cdot \nu} ]
) \quad \textrm{since } log \; x \leq x -1 \\
& = & 
\sum_{y \in \mathcal{Y}} (
k_{\lambda} [ \gamma \cdot \nu ]
+1 
- \sum_{x \in \mathcal{X}} (
p_{\lambda} (x) e^{ \sum^n_{i=1} \gamma_i \nu_i(x)
  \frac{\nu_\#(x)}{\nu_\#(x)}}
)
)\\
& = & 
\sum_{y \in \mathcal{Y}} (
k_{\lambda} [ \gamma \cdot \nu ]
+1 
- \sum_{x \in \mathcal{X}} (
p_{\lambda} (x) e^{ \sum^n_{i=1} \gamma_i \bar\nu_i(x) \nu_\#(x)}
)
)\\
& \geq &
\sum_{y \in \mathcal{Y}} (
k_{\lambda} [ \gamma \cdot \nu ]
+1 
- \sum_{x \in \mathcal{X}} (
p_{\lambda} (x) \sum^n_{i=1} \bar\nu_i(x) e^{  \gamma_i \nu_\#(x)}
)
) \\
& & \textrm{by Jensen's inequality}\\
& = & 
\sum_{y \in \mathcal{Y}} (
k_{\lambda} [ \gamma \cdot \nu ]
+1 
- p_{\lambda} [ \sum^n_{i=1} \bar\nu_ie^{  \gamma_i \nu_\#}
]
) \\
& = & 
A(\gamma + \lambda).
\qed
\end{eqnarray*}
\renewcommand{\qed}{}
\end{proof}

Lemma \ref{A0=0} shows that there is no estimated improvement in
log-\\ likelihood in the origin.

\begin{lemma} \label{A0=0}
$A(0+\lambda) = 0$. 
\end{lemma}

\begin{proof}
\[
A(0 + \lambda) 
= \sum_{y \in \mathcal{Y}} (
k_{\lambda} [ 0 \cdot \nu ]
+ 1 
- \sum_{x \in \mathcal{X}} p_{\lambda}(x) \sum^n_{i=1} \bar\nu_i (x)
e^0
)
=0.
\qed
\]
\renewcommand{\qed}{}
\end{proof}

Lemma \ref{dA=dL} shows that the critical points of interest are the
same for $A$ and $L$.

\begin{lemma} \label{dA=dL}
$ \left. \frac{d}{dt} \right|_{t=0} A(t\gamma + \lambda)
= \left. \frac{d}{dt} \right|_{t=0} L(t\gamma + \lambda)$.
\end{lemma}

\begin{proof}
\begin{eqnarray*}
\frac{d}{dt} A(t\gamma + \lambda) 
& = & \frac{d}{dt} 
\sum_{y \in \mathcal{Y}} (
k_{\lambda} [ t\gamma \cdot \nu ]
+1 
- \sum_{x \in \mathcal{X}} (
p_{\lambda} (x) \sum^n_{i=1} \bar\nu_i(x) e^{  t\gamma_i \nu_\#(x)}
)
) \\
& = & 
\sum_{y \in \mathcal{Y}} (
k_{\lambda} [ \gamma \cdot \nu ]
- \sum_{x \in \mathcal{X}} (
p_{\lambda} (x) \sum^n_{i=1} \frac{\nu_i(x)}{\nu_\#(x)} e^{  t\gamma_i
  \nu_\#(x)} \gamma_i \nu_\#(x)
)
) \\
& = &
\sum_{y \in \mathcal{Y}} (
k_{\lambda} [ \gamma \cdot \nu ]
- \sum_{x \in \mathcal{X}} (
p_{\lambda} (x) 
\sum^n_{i=1} \nu_i (x) \gamma_i e^{t \gamma_i \nu_\# (x)} 
)). \\
& & \\
\left. \frac{d}{dt} \right|_{t=0} A(t\gamma + \lambda)
& = & 
\sum_{y \in \mathcal{Y}} (
k_{\lambda} [ \gamma \cdot \nu ] 
- \sum_{x \in \mathcal{X}} ( 
p_{\lambda} (x) 
\sum^n_{i=1} \nu_i (x) \gamma_i e^0 
)) \\
& = &  
\sum_{y \in \mathcal{Y}} (
k_{\lambda} [ \gamma \cdot \nu ] 
- p_{\lambda} [ \gamma \cdot \nu ]
). \\
& & \\
\frac{d}{dt} L(t\gamma + \lambda) 
& = & 
\sum_{y \in \mathcal{Y}} (
\frac{d}{dt} log \sum_{x \in X(y)} p_{t\gamma + \lambda}(x)
) \\
& = & 
\sum_{y \in \mathcal{Y}} (
( \sum_{x \in X(y)} p_{t\gamma + \lambda}(x) )^{-1}
\frac{d}{dt}
\sum_{x \in X(y)}
e^{t\gamma \cdot \nu(x)} 
p_{\lambda}(x)
Z_{t\gamma + \lambda}^{-1}
)\\
& = & 
\sum_{y \in \mathcal{Y}} (
( \sum_{x \in X(y)} p_{t\gamma + \lambda}(x) )^{-1}
\sum_{x \in X(y)}
p_{\lambda}(x)  \\
& & 
( 
- e^{t\gamma \cdot \nu(x)} Z_{t\gamma + \lambda}^{-2} 
\sum_{x \in \mathcal{X}} e^{t\gamma \cdot \nu(x)} \gamma \cdot \nu(x) 
p_{\lambda}(x) \\
& & 
+
Z_{t\gamma + \lambda}^{-1} e^{t\gamma \cdot \nu(x)} \gamma \cdot
\nu(x)
))
\\
& = & 
\sum_{y \in \mathcal{Y}} (
- \sum_{x \in X(y)} p_{t\gamma + \lambda}(x)
p_{t\gamma + \lambda} [ \gamma \cdot \nu ]
( \sum_{x \in X(y)} p_{t\gamma + \lambda}(x))^{-1} \\
& & 
+ \sum_{x \in X(y)} p_{t\gamma + \lambda} [ \gamma \cdot \nu
]
( \sum_{x \in X(y)} p_{t\gamma + \lambda}(x))^{-1} 
) \\
& = & 
\sum_{y \in \mathcal{Y}} (
-  p_{t\gamma + \lambda} [ \gamma \cdot \nu
] 
+
k_{t\gamma + \lambda} [\gamma \cdot \nu ]
). \\
& & \\
\left. \frac{d}{dt} \right|_{t=0} L(t\gamma + \lambda)
& = & \sum_{y \in \mathcal{Y}} (
k_{\lambda} [\gamma \cdot \nu ]
- p_{\lambda} [ \gamma \cdot \nu
] 
).
\qed
\end{eqnarray*}
\renewcommand{\qed}{}
\end{proof}

One central result of this section is stated in Theorem
\ref{IncreasingLikelihood}. It  shows the hill-climbing
nature of the IM algorithm, i.e., the log-likelihood
$L$ is increasing on each iteration of the IM algorithm
except at fixed points of $\mathcal{M}$ or equivalently at critical
points of $L$. 

\newtheorem{theorem}[po]{Theorem}
\begin{theorem} \label{IncreasingLikelihood}
For all $\lambda \in \Lambda$:
$L(\mathcal{M}(\lambda) )\geq L(\lambda)$ with equality iff $\lambda$
is a fixed point of $\mathcal{M}$ or equivalently is a critical point of
$L$.
\end{theorem}

\begin{proof}
\begin{eqnarray*}
L(\mathcal{M}(\lambda)) - L(\lambda) 
& \geq & 
A(\mathcal{M}(\lambda)) \quad \textrm{by Lemma \ref{A<L-L}} \\
& \geq &
0 \quad \textrm{by Lemma \ref{A0=0} and definition of $\mathcal{M}$}.
\end{eqnarray*}
The equality $L(\mathcal{M}(\lambda)) = L(\lambda)$  holds iff
$\lambda$ is a fixed point of $\mathcal{M}$,
i.e., $\mathcal{M}(\lambda) = \hat\gamma + \lambda$ with
$\hat\gamma = 0$.
Furthermore, $\lambda$ is a fixed point of $\mathcal{M}$ iff 
$\hat\gamma = argmax_{\gamma \in \Rn} A(\gamma +
\lambda) = 0$, \\
$\iff \textrm{for all } \gamma \in \Rn : \hat t = argmax_{t \in
  \Reals } A(t\gamma +
\lambda) = 0$, \\
$\iff \textrm{for all } \gamma \in \Rn: \left. \frac{d}{dt} \right|_{t=0} A(t
\gamma + \lambda) = 0$, \\
$\iff \textrm{for all }\gamma \in \Rn: \left. \frac{d}{dt} \right|_{t=0}
L(t\gamma + \lambda) = 0 $, by Lemma \ref{dA=dL} \\
$\iff  \lambda \textrm{ is a critical point of $L$}.$
\end{proof}

Corollary \ref{MaximumLikelihoodEstimates} implies that a maximum
likelihood estimate is a fixed point of the mapping $\mathcal{M}$.

\newtheorem{corollary}[po]{Corollary}
\begin{corollary} \label{MaximumLikelihoodEstimates}
Let $\lambda^\ast = argmax_{\lambda \in \Lambda} L(\lambda)$.
Then $\lambda^\ast$ is a fixed point of $\mathcal{M}$.
\end{corollary}

Proposition \ref{convergence} discusses the convergence properties of
the algorithm. As with each application of the EM algorithm, we can
show convergence of the IM algorithm to critical points of the
incomplete-data log-likelihood function $L$. This means that the
limiting parameter value can occur at a local, not only at a global 
maximum of $L$. This chaotic behaviour of the algorithm, i.e., the
dependence of convergence on initial parameter values, must be treated
as an empirical matter.

\begin{po}\label{convergence}
Let $\{ \lambda^{(k)} \}$ be a sequence in $\Lambda$
determined by the IM Algorithm.
Then all limit points of $\{ \lambda^{(k)} \}$ are fixed points of
$\mathcal{M}$ or equivalently are critical points of $L$.
\end{po}

\begin{proof}
Let $\{ \lambda^{(k_n)} \} $ be a subsequence of
$\{ \lambda^{(k)} \}$ converging to $\bar\lambda$.
Then for all $\gamma \in
\Rn$:
\begin{eqnarray*}
A(\gamma + \lambda^{(k_n)} ) 
& \leq &
A(\hat\gamma^{(k_n)} + \lambda^{(k_n)} ) \quad \textrm{by definition
  of $\mathcal{M}$} \\
& \leq &
L(\hat\gamma^{(k_n)} + \lambda^{(k_n)}) - L(\lambda^{(k_n)}) \quad
\textrm{by Lemma \ref{A<L-L}} \\
& = &  
L( \lambda^{(k_n + 1)}) - L(\lambda^{(k_n)}) \quad \textrm{by
  definition of IM} \\
& \leq &
L( \lambda^{(k_{n+1})}) - L(\lambda^{(k_n)})
\end{eqnarray*}
and in limit as $n \rightarrow \infty$ for continuous $A$ and $L$:
$A(\gamma + \bar\lambda) \leq L(\bar\lambda) -  L(\bar\lambda) = 0$.
Thus $\gamma = 0$ is a maximum of $A(\gamma + \bar\lambda)$,
using Lemma \ref{A0=0},
and $\bar\lambda$ is a fixed point of $\mathcal{M}$.
Furthermore, $\left. \frac{d}{dt} \right|_{t=0} A(t\gamma +
\bar\lambda) = \left. \frac{d}{dt} \right|_{t=0} L(t\gamma +
\bar\lambda) = 0$,
using Lemma \ref{dA=dL},
and $\bar\lambda$ is a critical point of
$L$. 
\end{proof}

\subsection{Property Selection from Incomplete Data}
\label{PropertySelection}

For the preceding task of parameter estimation we assumed a vector of
properties to be given. However, exhaustive sets of properties can get
unmanageably large for most applications. Let us consider the
application of probabilistic CLP: One possible definition of properties
of proof trees is as subtrees of proof trees. If we want to be as flexible as
possible in the definition of subtree-properties and define a subtree
of a proof tree to be an arbitrary subgraph of a proof tree, then the
number of subtrees will grow exponentially in the number of
proof tree nodes. Clearly, the set of candidate properties must be
restricted by some quality measure.

Property selection addresses two general issues.
First, selecting prominent properties out of a set of possible properties
can be seen as inducing a proper model that captures only the salient
properties of the training data. This is one of the main tasks of
statistical machine learning. Second, compact models will disallow
overfitting the training data as could be done with models with one parameter per
training element. Instead, a proper model will allow
generalizations to new data and temper the overtraining problem.

Depending on the definition of properties (for the CLP
application, e.g., as connected subgraphs of proof
trees s.t.\ each node of a subgraph has either zero descendants or the
same number of descendants as the corresponding node of the
supergraph and the node sets of the subgraphs do not intersect)
and the definition of a procedure to incrementally construct
properties (e.g., by selecting from an initial set of query-node
properties and from properties built by performing  one-step
resolutions at terminal nodes of subtree-properties of the model),
we start from a set of candidate properties for a log-linear model.

But what should the above-mentioned quality measure be? We
could take as the measure the improvement in log-likelihood
when extending a model $p_\lambda$ based upon weighted property function
$\phi= \lambda \cdot \nu$
by a single candidate property\footnote{In the following we will refer
  to the property corresponding to property function $c$ as the
  ``property $c$''.} $c$ with parameter $\alpha$
to a model $p_{\alpha + \lambda}$ based on extended property function
$\alpha c + \phi$.
In its basic form this quality measure would require a calculation of
maximum likelihood estimates of extended models via the IM
algorithm for each candidate property. Clearly, this is not feasible
for models with large parameter spaces.
Following \citeN{Della-Pietra:95} or \citeN{Berger:96},
we could  instead approximate the improvement due to adding a single
property by adjusting only the parameter of this candidate and holding
all other parameters of the model fixed. This would make the property
selection algorithm practical but also greedy. Unfortunately, in constrast to this
approach, we cannot directly maximize the gain of adding property $c$
as a function of parameter $\alpha$ since the incomplete-data log-likelihood
$L$ is not concave in the parameters.
However, we can define an auxiliary function similar to the one used
in parameter estimation to express an approximate gain as a
conservative estimate of the log-likelihood difference.
A possible definition of an approximate gain 
can be derived from an
instantiation of the auxiliary function $A$ of
Sect. \ref{ParameterEstimation} to $A(\alpha + \lambda)$,
denoting the extension of a log-linear model $p_\lambda$ with
property-function vector $\nu$ by a single property $c$ with
log-parameter $\alpha$.

\begin{de}
Let $\phi = \lambda \cdot \nu$ be a weighted property function,
$c$ be a candidate property,
and $\alpha \in \Reals$ the log-parameter corresponding
to $c$.
Then the approximate gain $G_c(\alpha + \lambda)$
of adding candidate property $c$ with parameter value $\alpha$ to the
log-linear model $p_\lambda$ is defined s.t. 
\begin{center}
$ G_c(\alpha + \lambda) 
= \sum_{y \in \mathcal{Y}} (
1 + 
k_\lambda [\alpha c] - 
p_\lambda[e^{\alpha c}] )$
\end{center}
where 
$k_{\lambda}(x) = \frac{p_{\lambda}(x)}{\sum_{x \in X(y)}
 p_{\lambda}(x)}, \;
p_\lambda(x) = Z_\phi^{-1} e^{\phi(x)}p_0(x)$.
\end{de}

For this function similar properties hold as for the auxiliary function
$A$ of Sect. \ref{ParameterEstimation}.
Since $G_c$ is strictly concave in the parameters, we can maximize it
directly and uniquely as a function of $\alpha$.

\begin{po}
For each $\lambda \in \Lambda$, $\alpha \in \Reals$:
$G_c (\alpha + \lambda)$ takes its maximum as a
function of $\alpha$ at the unique point $\hat \alpha$ 
satisfying
\begin{center}
$\sum_{y \in \mathcal{Y}}
k_\lambda [c] 
= 
\sum_{y \in \mathcal{Y}}
p_\lambda [ c \: e^{\hat\alpha c} ]$.
\end{center}
\end{po}

\begin{proof}
\begin{quote}
$\frac{\partial}{\partial \alpha} G_c(\alpha + \lambda)
= \sum_{y \in \mathcal{Y}} (
k_\lambda [ c] - 
p_\lambda [c \: e^{ \alpha c}] )$, \\
$\frac{\partial^2}{\partial \alpha^2} 
 G_c(\alpha + \lambda) 
= - \sum_{y \in \mathcal{Y}} p_\lambda [c^2 e^{\alpha c}] < 0.$
\qed
\end{quote}
\renewcommand{\qed}{}
\end{proof}

Property selection then will incorporate that property out of the set
of candidates that gives greatest improvement to the model at the property's
best adjusted parameter value. Since we are interested only in
relative, not absolute gains, a single, non-iterative maximization of
the approximate gain will suffice to choose from the candidates.

\begin{de}[property selection]
Let $C$ be a set of candidate properties,
$c \in C$ be a candidate property with log-parameter
$\alpha \in \Reals$,
and $G_c (\lambda) = max_\alpha G_c (\alpha + \lambda)$
the maximal approximate gain that property $c$ can give to model
$p_\lambda$.
Then $c$ is selected in a property selection step for model
$p_\lambda$ if $c = argmax_{c' \in C} G_{c'}(\lambda)$.
\end{de}

\subsection{Summary}

The combined incomplete-data induction algorithm for log-linear models
can be summarized as follows.

{\em
\begin{description}
\item[\textbf{Input}] Initial model $p_0$, incomplete data set
  $\mathcal{Y}$.

\item[\textbf{Output}] Log-linear model $p^\ast$ on complete data set
$\mathcal{X} = \bigcup_{y \in \mathcal{Y}} X(y)$
with selected property function vector $\nu^\ast$ and 
log-parameter vector $\lambda^\ast = argmax_{\lambda \in \Lambda}
L(\lambda) $ 
where $\Lambda = \{ \lambda | \; p_\lambda$ 
is a log-linear model on $\mathcal{X}$ 
based on $p_0$, $\nu^\ast$ and $\lambda \in \Rn \} $.

\item[\textbf{Algorithm}] 
\begin{enumerate}

\item $p^{(0)} = p_0$,

\item Property selection: For each candidate property $c \in C^{(n)}$,
compute the gain $G_c(\lambda^{(n)}) = max_{\alpha \in \Reals} 
 G_c(\alpha + \lambda^{(n)})$
and select property $\hat c = argmax_{c \in C^{(n)}}
G_c(\lambda^{(n)})$.

\item Parameter estimation: Compute the maximum likelihood parameter
  value $\hat \lambda = argmax_{\lambda
    \in \Lambda} L(\lambda)$ 
where $\Lambda = \{ \lambda | \; p_\lambda (x) $
is a log-linear distribution on $\mathcal{X}$
with initial model $p_0$,
property function vector $\hat \nu = \nu^{(n)} \cup \hat c$, 
and $\lambda \in \Rn \} $.

\item Set $p^{(n+1)} = p_{\hat\lambda \cdot \hat\nu}$, 
$n = n+1$, go to $2$.

\end{enumerate}

\end{description}}

Returning to the sample program of Fig. \ref{program}, we can find a
simple log-linear reformulation of the probabilistic CLP model as
follows. In order to distinguish between the possible proof trees of
Fig. \ref{prooftrees} it is sufficient to define a single property
referring to the variable binding either to $a$ or to $b$. Taking a
parameter value of $log \: 2$ for a single property involving the
variable binding to $a$ will yield the desired probability
distribution $p(x_1) = 2/3$, $p(x_2) = 1/3$ and incomplete-data
log-likelihood $L = .148$.
The same result is obtained by taking a parameter value of $log \:1/2$
for a single property involving the variable binding to $b$. 
All other properties will be unable to distinguish between proof
trees $x_1$ and $x_2$ and thus give a uniform distribution over the proof
trees and log-likelihood $L= .125$.

\section{Approximation Methods}
\label{Approximation}

With the algorithms and proofs of the preceding section at hand,
induction of log-linear models from incomplete data reduces to a calculation of
expectations of simple functions. This calculation can be done by an
explicit summation over the configuration space only for probabilistic
processing models with a small, finite set of possible analyses.
In case of large or infinite configuration spaces and complex
parameter spaces these expectations can get intractable both analytically and
numerically. Here approximation methods have to be used.

Following \citeN{Della-Pietra:95} and \citeN{AbneySAVG:96},
we can use a combination of the approximation techniques of Newton's
method and Monte Carlo methods.
In order to give a self-contained recipe for inducing log-linear
models from incomplete data, we will make the proposed use of these
methods explicit in the following.

Newton's method is a technique to approximate the solution $\alpha$ of an
equation $f(\alpha) = 0$ by using a sequence of linearizations of
$f$. At each step the intersection of the tangent to
$f$ at $\alpha_t$ with the $\alpha$-axis is taken, yielding an
improved estimate $\alpha_{t+1}$. The iteration formulae to approach
the solution up to a desired accuracy are defined as follows:

{\em
\begin{center}
$\alpha_{t+1} = \alpha_t - \frac{f(\alpha_t)}
{f'(\alpha_t)}$ where $f'(\alpha_t)$ is the derivative of $f$ at $\alpha_t$.
\end{center}}

This method directly suits our application when we replace
$f(\alpha)$ by the first derivative of the auxiliary function $A$, $\frac{\partial}{\partial \gamma_i} A(\gamma +
\lambda)$, in case of parameter estimation, and by the first derivative
of the approximate gain $G_c$, $\frac{\partial}{\partial \alpha}
G_c(\alpha + \lambda)$, in case of property selection. Newton's method
usually converges rapidly for such functions.

The expectations expressed in the such defined Newton formulae then
can be estimated by Monte Carlo methods. A Monte Carlo technique
applicable to our problem is the Metropolis-Hastings method. The
strategy behind this method is to generate a random sample from a
target distribution $p$ via choosing a nominating matrix $p'$ from
which sampling is easy and performing a Bernoulli trial with parameter
$\alpha$ to determine whether to accept or reject the nominated
sample point. That means, this method converts a sampler for $p'$
into a sampler for $p$ via the evaluation matrix $\alpha$.
For our application, we can take as nominating matrix for each query
$y \in \mathcal{Y}$ a stochastic context-free CLP model
$p(x; \pi)$ on $X(y)$  as defined in Sect. \ref{Baum}. From this
stochastic derivation model sampling is easy and can be converted by a
standard evaluation matrix to sampling from the desired log-linear
distribution $p_\lambda(x)$ on $X(y)$. 
More formally, it can be shown that the  distribution of the sampled
random variables  $X_i$ will converge to the target distribution
$p_\lambda$ as $i\rightarrow \infty$, i.e., we have:
{\em
\begin{center}
$lim_{i \rightarrow \infty} P(X_i = x) = p_\lambda(x)$
for all $x \in X(y)$.
\end{center}}
Following standard textbooks such as \citeN{Fishman:96}, an
application of the Metropolis-Hastings algorithm to our problem is as follows.

{\em
\begin{description}

\item[\textbf{Input}] initial state $x_0 \in X(y)$, \\
nominating matrix $p' = p(x;\pi)$ on $X(y)$,\\
log-linear distribution $p = p_\lambda(x)$ on $X(y)$,\\
evaluation matrix $\alpha_{x,z} = 
\left\{ \begin{array}{ll}
1 & if \; p(x) p'(z) \leq p(z) p'(x) \\
\frac{p(z) p'(x)} {p(x) p'(z)} & if \;
p(x) p'(z) > p(z) p'(x) 
\end{array}
\right.$, \\
terminal number of steps $k$.

\item[\textbf{Output}] random sample $X_0, \ldots, X_k$
from $p_\lambda$ on $X(y)$.

\item[\textbf{Algorithm}] \mbox{} 
\begin{tabbing}
$X_0 := x_0$, \\
$i := 1$ ,\\
While \=  $i \leq k$ \\
\> $x := X_{i-1}$, \\
\> Randomly generate $z$ from $p'$, \\
\> If \= $z = X_{i-1}$ , then $X_i := X_{i-1}$, \\
\> \> Else evaluate $\alpha_{x,z}$, \\
\> \> Randomly generate $u$ from uniform distribution on $[0,1]$, \\
\> \> If \= $u \leq \alpha_{x,z}$ , then $X_i := z$ , \\
\> \> \> Else $X_i := X_{i-1}$, \\
\> $i:= i + 1$, \\
return $X_0, \ldots, X_k$.
\end{tabbing}

\end{description}}

In general, a proper random sample  from a probability distribution $p$
allows the estimation of expectations of functions $f$ with respect to $p$
directly from the sample points $X_i$, i.e., we have:
\begin{center}
$lim_{K \rightarrow \infty} \frac{1}{K} \sum_{i=1}^K f(X_i) =
\sum_x f(x) p(x)$.
\end{center}
For our application taking a random sample $\tilde X(y)$
from $p_\lambda$ on $X(y)$ for a query $y \in \mathcal{Y}$
will allow us to calculate expectations of functions with respect to the
distribution $p_\lambda$ on $X(y)$. A combination of $i$-ary random
samples $\tilde X(y)$ from $p_\lambda$ on $X(y)$
for queries $y \in \mathcal{Y}$ will yield a combined random sample
$\tilde \mathcal{X} = \bigcup_{y \in \mathcal{Y}} \tilde X(y)$
from $p_\lambda$ on
$\mathcal{X} = \bigcup_{y \in \mathcal{Y}} X(y)$.
From this random sample we can then estimate expectations of functions
with respect to a distribution of $p_\lambda$ on $\mathcal{X}$.

Returning to the estimation of the expectations involved in our
induction formulae, we note that we can use the same random sample
from $p^{(n)}$ for each iteration of Newton's method in
estimating the gain $G_c(\lambda^{(n)})$ for each candidate property
$c \in C^{(n)}$ simultaneously.
After adding a selected property $\hat c$ to the model, we can again
use a single random sample from the extended model for the estimation of
the maximum likelihood parameter values via Newton's method for each
property in parallel.
This means that we can build up hash-tables counting up how many times
each property takes on which value. 
Let $\mathcal{Y}$ be an incomplete data sample of size $N$, 
$\tilde X(y)$ be a complete data sample of size $M$
for $y$, and $\tilde \mathcal{X}$ be a combined complete data sample
of size $L$. Then the relevant hash-tables can be defined as follows:

{\em
\begin{enumerate}
\item $S_{c,v} = \sum_{\tilde x \in \tilde \mathcal{X}} 
[ \! [ c(\tilde x) = v ] \! ] $
is the number of times property function $c$ takes value $v$ in
combined random sample $\tilde \mathcal{X}$,
\item $T_{y,c,v}=\sum_{\tilde x \in \tilde X(y)} 
[ \! [ c(\tilde x) = v ] \! ] $
is the number of times property function $c$ takes value $v$ in
random sample $\tilde X(y)$, 
\item $U_{i,m}=\sum_{\tilde x \in \tilde \mathcal{X} | \;
  \nu_\# (\tilde x) = m} \nu_i(\tilde x)$ 
is the number of times property $\chi_i$ appears in combined random
sample $\tilde \mathcal{X}$ when there is a total number of $m$
property instances for each sample point. 
\end{enumerate}}
Furthermore, it will be convenient to define the following variables:

\begin{quote}
$s_r(\alpha,c) = \sum_v S_{c,v} e^{\alpha v} v^r$, \\
$t_y(c) = \frac{1}{M} \sum_v T_{y,c,v} v$, \\
$u_r(\alpha, i) = \sum_m U_{i,m} e^{\alpha m} m^r$.
\end{quote}
The expectations involved in Newton's formulae for the property
selection task can then be approximated by random sample counts as follows:

\begin{eqnarray*}
\alpha_{t+1}
& = & 
\alpha_t +
\frac{ \frac{\partial}{\partial \alpha_t}
G_c(\alpha_t + \lambda) }
{ \frac{\partial^2}{\partial \alpha_t^2}
G_c(\alpha_t + \lambda) }
\\
& = & 
\alpha_t +
\frac{\sum_y k_\lambda [c] - N p_\lambda
[ c \: e^{\alpha_t c} ] }
{N p_\lambda [c^2 e^{\alpha_t c} ]}
\\
 &  \approx & 
\alpha_t +
\frac{\sum_y t_y(c) - \frac{N}{L} s_1(\alpha_t,c)}
{ \frac{N}{L} s_2(\alpha_t,c)}.
\end{eqnarray*}
Similar estimation formulae can be obtained for the task of parameter estimation:

\begin{eqnarray*}
\alpha_{t+1}
& = & 
\alpha_t +
\frac{ \frac{\partial}{\partial \alpha_t}
A(\gamma + \lambda) }
{ \frac{\partial^2}{\partial \alpha_t^2}
A(\gamma + \lambda) }
\\
& = & 
\alpha_t +
\frac{\sum_y k_\lambda [\nu_i] - N p_\lambda [\nu_i e^{\alpha_t
  \nu_\#}]}
{N p_\lambda [\nu_i \nu_\# e^{\alpha_t \nu_\#} ]}
\\
 &  \approx & 
\alpha_t +
\frac{\sum_y t_y(\nu_i) - \frac{N}{L} u_0(\alpha_t, i)}
{ \frac{N}{L} u_1(\alpha_t, i)}.
\end{eqnarray*}

\section{Search Methods}
\label{Search}

The induction and approximation techniques of the preceding sections
provide the means to induce a proper probability distribution over
analyses of a log-linear probabilistic processing model from
unanalyzed input data. In case of ambiguity, this allows us to
distinguish between analyses according to a well-defined and practical quality
measure. However, if we are interested only in the best analysis of a
given input, so far a ranking of analyses requires a listing of
analyses in order to choose the best one. Clearly, it would be nice to
have search techniques like Viterbi's algorithm
\cite{Viterbi:67}, which works well for probabilisitic processing
models based on context-free stochastic derivation processes.

Viterbi's algorithm is built upon a table of derivation states, called
a chart, describing different pending derivations. During
derivation, each state must keep track of the most probable path of
states leading towards it. When the final state is reached, the maximum
probability derivation can be recovered by tracing back the path of
the best predecessor states. Different specifications of the algorithm
depend on the chosen parsing strategy and the underlying probabilistic
model.

In the following we will sketch one possibility to transfer these
ideas to a method of probabilistic parsing in the area of CLP. For
this aim we rely on the well-known parsing algorithm of Earley
deduction. This technique provides the necessary chart
structure accompanied with a simple parsing strategy. Depending on the
specific definition of the property vector  in the underlying
log-linear CLP model, different definitions of the propagation of
probabilities during the parsing process are possible. Since the
property vector is considered to be an open parameter in our setting,
we will not present a definitive solution to this problem but only
give some rules of thumb how to proceed for some general examples.

Earley deduction was introduced by \citeN{Pereira:83} as a
generalization of Earley's context-free parsing algorithm (see
\citeN{Aho:72}) to a parsing algorithm for definite
clause grammars. Extensions of this method in the general setting of the
CLP scheme of \citeN{HuS:88} have been presented, e.g.,
by \citeN{Doerre:93a} and \citeN{Doerre:95}. 
The basic concepts of Earley deduction for CLP can be described as follows:
Earley deduction works on two sets of definite
clauses, the set of program clauses $\mathcal{P}$ and the set of
derived clauses constituting the chart $\mathcal{C}$. An active item
corresponds to a definite clause with at least one relational atom on
its righthandside, i.e., to a non-unit clause. Passive items
correspond to clauses whose righthandsides consist only of an
\La-constraint, i.e., to unit-clauses. The input to the algorithm
consists of a set of program clauses \Po and a query $G$. The 
content of the chart $\mathcal{C}$ initially consists of $G$ and is
continually added to by the following inference rules:

{\em
\begin{quote}
\textbf{Prediction:} \\
\mbox{} \\
$c_1 = (H_1 \leftarrow B_1) \: \in \mathcal{C}$ \\
$c_2 = (H_2 \leftarrow B_2) \: \in \Po$ \\
\put(0,0){\line(1,0){110}} \\
$c_3 = (C \leftarrow B_2' \cup \phi) \: \in \mathcal{C}$\\
\mbox{}\\
where $c_1$ is non-unit, $c_2$ is unit or non-unit, $C$ is the
selected literal in $B_1$, $\phi$ is the \La-constraint in $B_1$, and
 there exists a variant $c_2' = (C \leftarrow B_2')$
of $c_2$ 
s.t. $\Xsf{V}(c_1) \cap \Xsf{V}(B_2')
\subseteq \Xsf{V}(C)$.
\end{quote}

\vspace{1.5ex}

\begin{quote}
\textbf{Completion:} \\
\mbox{}\\
$c_1 = (H_1 \leftarrow B_1) \: \in \mathcal{C}$ \\
$c_2 = (H_2 \leftarrow B_2) \: \in \mathcal{C}$\\
\put(0,0){\line(1,0){140}} \\
$c_3 = (H_1 \leftarrow ( B_1 \setminus C) \cup B_2') \: \in
\mathcal{C}$ \\
\mbox{} \\
where $c_1$ is non-unit, $c_2$ is unit, $C$ is the
selected literal in $B_1$, and
 there exists a variant $c_2' = (C \leftarrow B_2')$
of $c_2$ 
s.t. $\Xsf{V}(c_1) \cap \Xsf{V}(B_2')
\subseteq \Xsf{V}(C)$.
\end{quote}}

A probabilistic version of a context-free Earley parser was presented
in \citeN{Stolcke:93}. In this framework, during derivation each
completed state keeps track of the most probable path of states
contributing to it. The probability propagation is done recursively by
associating each predicted  state with the probability of the
corresponding rule and taking at each completion step the maximum of all products
of probabilities of two states from which the completed state is
derivable. When the final state is reached, the most probable analysis
easily can be retrieved by building up a tree in accordance with the
most probable path of states leading to the final completion.

If the property vector of a log-linear CLP model is defined
s.t.\ properties are identified with program clauses, then the above
model can be used also for probabilistic Earley deduction: 
During deduction, each predicted clause is associated with a weight
corresponding to the clause-property used in the prediction. For each
completed clause, the pair of clauses contributing with maximal
product of weights to the completion is recorded. Given a procedure to
construct a proof tree from a sequence of clauses linked by
prediction and completion, the highest weighted partial proof tree
corresponding to a completed clause can be constructed recursively and
uniquely from the highest weighted pair of clauses contributing to the
completion.

Unfortunately, weight propagation will get more complicated as we
allow more complicated properties in our underlying log-linear CLP
model. In case properties are identified with program clauses,
completion means complete reduction of selected atoms using
appropriate clauses. A numerical comparison between different ways of
arriving at the same completed state can be done at every completion
step. In contrast to this, if properties are allowed to be subtrees of
proof trees, completion means completely building up a subtree of a
proof tree during derivation. A numerical comparison between to ways
of ``completing'' the same subtree in the same completion state might
have to wait for several completion steps until the subtree is completely built
up. Considering the possibility of a backward construction of the most
probable proof tree in this setting, we cannot rely on an easy
recording of the most probable path of clauses leading to the final
completion state. Instead, in order to compare between the
weights of the partial derivations contributing to such a ``subtree-completion'', we
have to incrementally build up partial proof trees and check their
properties during derivation.

Let subtree-properties be defined as follows: A subtree of a
proof tree is a connected subgraph of a proof tree, each node of
a subgraph has either zero descendants or the same number of
descendants as the corresponding node of the supergraph, and the node
sets of every two subtrees in the set of properties do not
intersect. Then a simple recursive procedure to build up partial proof trees
from completed states can be defined as follows:

{\em
\begin{quote}

For each completed state $c_k$, for each pair of states $c_i, c_j$
from which $c_k$ is derivable by completion, the partial proof tree
$t_{ij}$ corresponding to the completion of state $c_k$
from states $c_i, c_j$ is constructed s.t. 
$t_{ij} = $

\begin{enumerate}

\item
$\begin{array}{c}
t_i \\
\oplus \\
t_j
\end{array} $,
if $c_i,c_j$ are completed states with trees $t_i, t_j$,

\hspace{5ex}
and
{\tiny
$\begin{array}{ccc}
 & & \vdots \\
 & & A\\
t_1& & | \\
\oplus & = & B \cup C \\
t_2 & & | \\
 & & D \\
 & & \vdots
\end{array} $}
if
{\tiny
$\begin{array}{ccc}
& & \vdots \\
t_1 & = & A \\
& & | \\
& & B
\end{array}$},
{\tiny
$\begin{array}{ccc}
& & C \\
& & | \\
t_2 & = & D \\
& & \vdots
\end{array}$},

\item 
$\begin{array}{c}
t_i \\
\otimes \\
t_j
\end{array}$,
if $c_i$ is a predicted state $(E \leftarrow F)$ with tree
{\tiny
$\begin{array}{ccc}
& & E \\
t_i & = & | \\
& & F
\end{array}$},

\hspace{5ex}
$c_j$ is a completed state with tree $t_j$, 

\hspace{5ex}
and
{\tiny
$\begin{array}{ccc}
 & & \vdots \\
 & & A \\
t_1& & | \\
\otimes & = & B \\
t_2 & & | \\
 & & B \setminus C \cup D\\
 & & \vdots
\end{array} $}
if
{\tiny
$\begin{array}{ccc}
& & \vdots \\
t_1 & = & A \\
& & | \\
& & B
\end{array}$},
{\tiny
$\begin{array}{ccc}
& & C \\
& & | \\
t_2 & = & D \\
& & \vdots
\end{array}$}, 

\item
$\begin{array}{c}
t_i \\
| \\
t_j
\end{array}$,
if $c_i$ is a predicted state $(E \leftarrow F)$
with tree
{\tiny
$\begin{array}{ccc}
& & E \\
t_i & = & | \\
& & F
\end{array}$},

\hspace{5ex}
$c_j$ is a predicted state including
\La-constraint $\phi$ 

\hspace{5ex} 
and with tree $t_j = \phi$.

\end{enumerate}

\end{quote}
}

During derivation, for each ``property-completion'' at some completed state
$c_k$, the variable $t_k$ denoting the partial proof tree
corresponding to $c_k$ is instantiated to the most probable partial
proof tree $t_{ij}$ which can be built from all states $c_i, c_j$ contributing
to the completion of $c_k$:

{\em
\begin{quote}
Let $p_\lambda$ be a log-linear distribution on the set $\mathcal{X}$
of proof trees of a constraint logic program \Po with property vector
$\chi$ and property function vector $\nu$.
Then for each completed state $c_k$,
for each property $\chi_n \in \chi$,
for each partial proof tree $t_{ij}$ constructable for $c_k$ from
trees $t_i, t_j$
s.t.\ $\nu_n(t_{ij}) > \nu_n(t_i) + \nu_n(t_j)$, 
set $t_k = argmax_{t_{ij}}\; p_\lambda (t_{ij})$.
\end{quote}
}

For the above definition of subtree-properties, this procedure
guarantees that the most probable proof tree is built up during
derivation. The possible savings in computational complexity induced
by this procedure clearly depend on the size of the
subtree-properties to be worked out during derivation. However, if
subtree-properties are allowed to be overlapping or disconnected
subgraphs of proof trees, then the above dynamic programming approach
is no longer applicable. In this case either exhaustive search or
approximation methods are required.

\section{Conclusion} 
\label{Hugh}

We presented a log-linear probability model for probabilistic CLP. On
top of this model we defined an algorithm to estimate the parameters
and to select the properties of log-linear models from incomplete
data. This algorithm is an extension of the iterative scaling algorithm of
\citeN{Della-Pietra:95} adjusted to incomplete data. The algorithm
applies to log-linear models in general and is accompanied with
suitable approximation methods when applied to large data
spaces. Furthermore, we presented an approach to search for most
probable analyses of the probabilistic CLP model. This can be useful
for the ambiguity resolution problem in natural language processing
applications.

Compared with Abney's approach to a log-linear model
for stochastic attribute-value grammars, our approach adds the
important aspect of incomplete data to the parameter estimation and
property selection problem. Furthermore, we investigate the problem of
searching for best analyses which is not addressed by
\citeN{AbneySAVG:96}.

The expressive power of log-linear models even allows us to couch other
approaches to probabilisitic processing beyond context-freeness
in terms of this framework. Statistical decision trees as used in
the probabilistic parsing model of \citeN{Magerman:94} can be cast in
the log-linear framework by encoding the questions building up a
decision tree as binary-valued, disjoint property functions. Property selection
then can be seen as closely related to growing a decision tree and
iterative maximization can be seen as maximum likelihood estimation
for such defined decision trees.
However, in contrast to the algorithms used by \citeN{Magerman:94},
which require large samples of complete data, our approach allows
induction of the probabilistic model from incomplete data.

A similar statement can be made for the probabilistic tree
substitution model of \citeN{Bod:95}. This approach can be couched as
a log-linear model employing all subtrees of a tree bank, which is annotated
according to some grammar framework, as properties of the model. Again,
Bod's approach relies on hand-analyzed data and does not
allow to estimate the probabilistic model from unanalyzed
input. Furthermore, this approach does not provide a means to
automatically select subtree-properties from the exponentially many
candidates. 

Clearly, our model of probabilistic CLP is not the last word on
probabilistic processing beyond context-freeness.
As mentioned above, log-linear models are closely related to other
probabilistic models such as random fields \cite{Geman:90}, graphical
networks \cite{Pearl:88} or neural networks \cite{Ackley:85}.
Future work should exploit this resemblance in order to learn from
related techniques to induce, approximate or search in log-linear
probability models. Furthermore, the possibilities of our powerful processing
model shall be applied to natural language processing tasks other than
ambiguity resolution.

\bibliographystyle{chicago}

\begin{thebibliography}{}

\bibitem[\protect\citeauthoryear{Abney}{Abney}{1996}]{AbneySAVG:96}
Abney, S. (1996).
\newblock Stochastic attribute-value grammars.
\newblock Unpublished manuscript, University of T\"ubingen. To appear in
  Computational Linguistics.

\bibitem[\protect\citeauthoryear{Ackley and Hinton}{Ackley and
  Hinton}{1985}]{Ackley:85}
Ackley, D.~H. and G.~E. Hinton (1985).
\newblock A learning algorithm for {Boltzmann} machines.
\newblock {\em Cognitive Science\/}~{\em 9}, 147--169.

\bibitem[\protect\citeauthoryear{Aho and Ullman}{Aho and Ullman}{1972}]{Aho:72}
Aho, A.~V. and J.~D. Ullman (1972).
\newblock {\em The Theory of Parsing, Translation and Compiling}, Volume I:
  Parsing.
\newblock NJ: Prentice-Hall.

\bibitem[\protect\citeauthoryear{Baker}{Baker}{1979}]{Baker:79}
Baker, J. (1979).
\newblock Trainable grammars for speech recognition.
\newblock In D.~Klatt and J.~Wolf (Eds.), {\em Speech Communication Papers for
  the 97th Meeting of the Acoustical Society of America}, pp.\  547--550.

\bibitem[\protect\citeauthoryear{Baum}{Baum}{1972}]{Baum:72}
Baum, L.~E. (1972).
\newblock An inequality and associated maximization technique in statistical
  estimation for probabilistic functions of {M}arkov processes.
\newblock {\em Inequalities\/}~{\em III}, 1--8.

\bibitem[\protect\citeauthoryear{Baum and Eagon}{Baum and
  Eagon}{1967}]{Baum:67}
Baum, L.~E. and J.~A. Eagon (1967).
\newblock An inequality with applications to statistical estimation for
  probabilistic functions of {M}arkov processes and to a model for ecology.
\newblock {\em Bulletin of the American Mathematical Society\/}~{\em 73\/}(1),
  360--363.

\bibitem[\protect\citeauthoryear{Baum, Petrie, Soules, and Weiss}{Baum
  et~al.}{1970}]{Baum:70}
Baum, L.~E., T.~Petrie, G.~Soules, and N.~Weiss (1970).
\newblock A maximization technique occuring in the statistical analysis of
  probabilistic functions of {M}arkov chains.
\newblock {\em The Annals of Mathematical Statistics\/}~{\em 41\/}(1),
  164--171.

\bibitem[\protect\citeauthoryear{Beeferman, Berger, and Lafferty}{Beeferman
  et~al.}{1997a}]{LaffertyACL:97}
Beeferman, D., A.~Berger, and J.~Lafferty (1997a).
\newblock A model of lexical attraction and repulsion.
\newblock In {\em Proceedings of the 35th Annual Meeting of the Association for
  Computational Linguistics}, Madrid, Spain.

\bibitem[\protect\citeauthoryear{Beeferman, Berger, and Lafferty}{Beeferman
  et~al.}{1997b}]{LaffertyEMNLP:97}
Beeferman, D., A.~Berger, and J.~Lafferty (1997b).
\newblock Text segmentation using exponential models.
\newblock In {\em Proceedings of the 2nd Conference on Empirical Methods in
  Natural Language Processing}.

\bibitem[\protect\citeauthoryear{Berger, {Della Pietra}, and {Della
  Pietra}}{Berger et~al.}{1996}]{Berger:96}
Berger, A.~L., V.~J. {Della Pietra}, and S.~A. {Della Pietra} (1996).
\newblock A maximum entropy approach to natural language processing.
\newblock {\em Computational Linguistics\/}~{\em 22\/}(1), 39--71.

\bibitem[\protect\citeauthoryear{Bod}{Bod}{1995}]{Bod:95}
Bod, R. (1995).
\newblock {\em Enriching Linguistics with Statistics: Performance Models of
  Natural Language}.
\newblock Ph.\ D. thesis, Institute for Logic, Language and Computation,
  Universiteit van Amsterdam.

\bibitem[\protect\citeauthoryear{Briscoe and Waegner}{Briscoe and
  Waegner}{1992}]{Briscoe:92}
Briscoe, T. and N.~Waegner (1992).
\newblock Robust stochastic parsing using the inside-outside algorithm.
\newblock In {\em Proceedings of AAAI92 Workshop on Probabilistically-Based
  Natural Language Processing Techniques}, San Jose, CA.

\bibitem[\protect\citeauthoryear{Carroll and Charniak}{Carroll and
  Charniak}{1992}]{Carroll:92}
Carroll, G. and E.~Charniak (1992).
\newblock Two experiments on learning probabilistic dependency grammars from
  corpora.
\newblock Technical Report RI 02912, Department of Computer Science, Brown
  University, Providence RI.

\bibitem[\protect\citeauthoryear{Darroch and Ratcliff}{Darroch and
  Ratcliff}{1972}]{Darroch:72}
Darroch, J. and D.~Ratcliff (1972).
\newblock Generalized iterative scaling for log-linear models.
\newblock {\em The Annals of Mathematical Statistics\/}~{\em 43\/}(5),
  1470--1480.

\bibitem[\protect\citeauthoryear{{Della Pietra}, {Della Pietra}, and
  Lafferty}{{Della Pietra} et~al.}{1995}]{Della-Pietra:95}
{Della Pietra}, S., V.~{Della Pietra}, and J.~Lafferty (1995).
\newblock Inducing features of random fields.
\newblock Technical Report CMU-CS-95-144, CMU.
\newblock Also appeared in {IEEE} Transactions on Pattern Analysis and Machine
  Intelligence 19(4), pp. 380-393, April 1997.

\bibitem[\protect\citeauthoryear{Dempster, Laird, and Rubin}{Dempster
  et~al.}{1977}]{Dempster:77}
Dempster, A.~P., N.~M. Laird, and D.~B. Rubin (1977).
\newblock Maximum likelihood from incomplete data via the {{\em EM}} algorithm.
\newblock {\em Journal of the Royal Statistical Society\/}~{\em 39\/}(B),
  1--38.

\bibitem[\protect\citeauthoryear{D\"orre}{D\"orre}{1993}]{Doerre:93a}
D\"orre, J. (1993).
\newblock Generalizing {Earley} deduction for constraint-based grammars.
\newblock In J.~D\"orre (Ed.), {\em Computational Aspects of Constraint-Based
  Linguistic Description I}, pp.\  25--41. DYANA-2 Deliverable R1.2.A.

\bibitem[\protect\citeauthoryear{D\"orre and Dorna}{D\"orre and
  Dorna}{1993}]{Doerre:93}
D\"orre, J. and M.~Dorna (1993).
\newblock {CUF} - a formalism for linguistic knowledge representation.
\newblock In J.~D\"orre (Ed.), {\em Computational Aspects of Constraint-Based
  Linguistic Description I}, pp.\  3--22. DYANA-2 Deliverable R1.2.A.

\bibitem[\protect\citeauthoryear{D\"orre and Johnson}{D\"orre and
  Johnson}{1995}]{Doerre:95}
D\"orre, J. and M.~Johnson (1995).
\newblock Memoization of coroutined constraints.
\newblock In {\em Proceedings of the 33rd Annual Meeting of the Association for
  Computational Linguistics}, pp.\  100--107.

\bibitem[\protect\citeauthoryear{Eisele}{Eisele}{1994}]{Eisele:94}
Eisele, A. (1994).
\newblock Towards probabilistic extensions of constraint-based grammars.
\newblock In J.~D\"orre (Ed.), {\em Computational Aspects of Constraint-Based
  Linguistic Description II}, pp.\  3--21. DYANA-2 Deliverable R1.2.B.

\bibitem[\protect\citeauthoryear{Fishman}{Fishman}{1996}]{Fishman:96}
Fishman, G.~S. (1996).
\newblock {\em Monte Carlo. Concepts, Algorithms and Applications}.
\newblock Berlin: Springer.

\bibitem[\protect\citeauthoryear{Geman}{Geman}{1990}]{Geman:90}
Geman, D. (1990).
\newblock Random fields and inverse problems in imaging.
\newblock In P.~L. Hennequin (Ed.), {\em {\' E}cole d'{\'E}t{\'e} de
  Probabilit{\'e}s de Saint-Flour {XVIII} - 1988}, pp.\  117--193. Berlin:
  Springer.
\newblock Lecture Notes in Mathematics, 1427.

\bibitem[\protect\citeauthoryear{Geman and Geman}{Geman and
  Geman}{1984}]{Geman:84}
Geman, S. and D.~Geman (1984).
\newblock Stochastic relaxation, {G}ibbs distributions, and the {B}ayesian
  restoration of images.
\newblock {\em {IEEE} Transactions on Pattern Analysis and Machine
  Intelligence\/}~{\em PAMI-6}, 721--741.

\bibitem[\protect\citeauthoryear{G\"otz}{G\"otz}{1995}]{Goetz:95}
G\"otz, T. (1995).
\newblock Compiling {HPSG} constraint grammars into logic programs.
\newblock In {\em Workshop on Computational Logic and Natural Language
  Processing}, Edinburgh.

\bibitem[\protect\citeauthoryear{H\"ohfeld and Smolka}{H\"ohfeld and
  Smolka}{1988}]{HuS:88}
H\"ohfeld, M. and G.~Smolka (1988).
\newblock Definite relations over constraint languages.
\newblock {LILOG} Report~53, IBM Deutschland, Stuttgart.

\bibitem[\protect\citeauthoryear{Jaffar and Lassez}{Jaffar and
  Lassez}{1986}]{JuL:86}
Jaffar, J. and J.-L. Lassez (1986).
\newblock Constraint logic programming.
\newblock Technical report~74, Department of Computer Science, Monash
  University.

\bibitem[\protect\citeauthoryear{Jaynes}{Jaynes}{1957}]{Jaynes:57}
Jaynes, E.~T. (1957).
\newblock Information theory and statistical mechanics.
\newblock {\em Physical Review\/}~{\em 106}, 620--630.

\bibitem[\protect\citeauthoryear{Jaynes}{Jaynes}{1983}]{Jaynes:83}
Jaynes, E.~T. (1983).
\newblock {\em Papers on Probability, Statistics and Statistical Physics}.
\newblock Dortrecht: D. Reidel.
\newblock ed. by R. D. Rosenkrantz.

\bibitem[\protect\citeauthoryear{Jelinek, Lafferty, and Mercer}{Jelinek
  et~al.}{1990}]{Jelinek:90}
Jelinek, F., J.~D. Lafferty, and R.~L. Mercer (1990).
\newblock Basic methods of probabilistic context free grammars.
\newblock Technical report, Continuous Speech Recognition Group IBM - T.J.
  Watson Research Center, Yorktown Heights, NY.

\bibitem[\protect\citeauthoryear{Lari and Young}{Lari and
  Young}{1990}]{Lari:90}
Lari, K. and S.~J. Young (1990).
\newblock The estimation of stochastic context-free grammars using the
  inside-outside algorithm.
\newblock {\em Computer Speech and Language\/}~{\em 4}, 35--56.

\bibitem[\protect\citeauthoryear{Lloyd}{Lloyd}{1987}]{Lloyd:87}
Lloyd, J.~W. (1987).
\newblock {\em Foundations of Logic Programming}.
\newblock Berlin: Springer.

\bibitem[\protect\citeauthoryear{Magerman}{Magerman}{1994}]{Magerman:94}
Magerman, D.~M. (1994).
\newblock {\em Natural Language Parsing as Statistical Pattern Recognition}.
\newblock Ph.\ D. thesis, Department of Computer Science, Stanford University.

\bibitem[\protect\citeauthoryear{Mark, Miller, Grenander, and Abney}{Mark
  et~al.}{1992}]{Mark:92}
Mark, K., M.~Miller, U.~Grenander, and S.~Abney (1992).
\newblock Parameter estimation for constrained context-free language models.
\newblock In {\em {DARPA} Speech and Natural Language Workshop}, Harriman, New
  York.

\bibitem[\protect\citeauthoryear{Miyata}{Miyata}{1996}]{Miyata:96}
Miyata, T. (1996).
\newblock {\em A Study on Inference Control in Natural Language Processing}.
\newblock Ph.\ D. thesis, Graduate School of the University of Tokyo, Tokyo,
  Japan.

\bibitem[\protect\citeauthoryear{Osborne and Briscoe}{Osborne and
  Briscoe}{1997}]{Osborne:97}
Osborne, M. and T.~Briscoe (1997).
\newblock Learning stochastic categorial grammars.
\newblock In {\em CoNLL97: Proceedings of the Workshop on Computational Natural
  Language Learning}, Madrid, Spain, pp.\  80--87.

\bibitem[\protect\citeauthoryear{Pearl}{Pearl}{1988}]{Pearl:88}
Pearl, J. (1988).
\newblock {\em Probabilistic Reasoning in Intelligent Systems: Networks of
  Plausible Inference}.
\newblock CA: Morgan Kaufmann.

\bibitem[\protect\citeauthoryear{Pereira and Schabes}{Pereira and
  Schabes}{1992}]{Pereira:92}
Pereira, F. and Y.~Schabes (1992).
\newblock Inside-outside reestimation from partially bracketed corpora.
\newblock In {\em Proceedings of the 30th Annual Meeting of the Association for
  Computational Linguistics}, Newark, Delaware, pp.\  128--135.

\bibitem[\protect\citeauthoryear{Pereira and Warren}{Pereira and
  Warren}{1983}]{Pereira:83}
Pereira, F. C.~N. and D.~H.~D. Warren (1983).
\newblock Parsing as deduction.
\newblock In {\em Proceedings of the 21st Annual Meeting of the Association for
  Computational Linguistics}, Boston, MA, pp.\  137--144.

\bibitem[\protect\citeauthoryear{Rabiner}{Rabiner}{1989}]{Rabiner:89}
Rabiner, L.~R. (1989).
\newblock A tutorial on hidden {M}arkov models and selected applications in
  speech recognition.
\newblock In {\em Proceedings of the IEEE, Vol. 77, No. 2}, pp.\  257--285.

\bibitem[\protect\citeauthoryear{Ratnaparkhi}{Ratnaparkhi}{1996}]{Ratnaparkhi:%
96}
Ratnaparkhi, A. (1996).
\newblock A maximum entropy model for part-of-speech tagging.
\newblock In {\em Proceedings of the 1st Conference on Empirical Methods in
  Natural Language Processing}.

\bibitem[\protect\citeauthoryear{Resnik}{Resnik}{1992}]{Resnik:92}
Resnik, P. (1992).
\newblock Probabilistic tree-adjoining grammars as a framework for statistical
  natural language processing.
\newblock In {\em Proceedings of COLING-92}, Nantes, pp.\  418--424.

\bibitem[\protect\citeauthoryear{Rosenfeld}{Rosenfeld}{1996}]{Rosenfeld:96}
Rosenfeld, R. (1996).
\newblock A maximum entropy approach to adaptive statistical language modeling.
\newblock {\em Computer, Speech and Language\/}~{\em 10}, 187--228.

\bibitem[\protect\citeauthoryear{Schabes}{Schabes}{1992}]{Schabes:92}
Schabes, Y. (1992).
\newblock Stochastic lexicalized tree-adjoining grammars.
\newblock In {\em Proceedings of COLING-92}, Nantes, pp.\  426--432.

\bibitem[\protect\citeauthoryear{Stolcke}{Stolcke}{1993}]{Stolcke:93}
Stolcke, A. (1993).
\newblock An efficient probabilistic context-free parsing algorithm that
  computes prefix probabilities.
\newblock Technical Report TR-93-065, International Computer Science Institute,
  Berkeley, CA.

\bibitem[\protect\citeauthoryear{Viterbi}{Viterbi}{1967}]{Viterbi:67}
Viterbi, A. (1967).
\newblock Error bounds for convolutional codes and an asymptotically optimum
  decoding algorithm.
\newblock {\em {IEEE} Transactions on Information Theory\/}~{\em IT-13},
  260--269.

\end{thebibliography}

\end{document}